\documentclass[journal]{IEEEtran}

\usepackage{soul}
\usepackage[table,xcdraw]{xcolor}

\newcommand{\new}[1]{#1}

\usepackage{lipsum}
\usepackage{amsmath}

\usepackage{physics}

\ifCLASSINFOpdf
  \usepackage[pdftex]{graphicx}

\else

\fi

\ifCLASSOPTIONcompsoc
 \usepackage[caption=false,font=normalsize,labelfont=sf,textfont=sf]{subfig}
\else
 \usepackage[caption=false,font=footnotesize]{subfig}
\fi

\begin{document}

\title{FAVbot: An Autonomous Target Tracking Micro-Robot with Frequency Actuation Control}

\author{Zhijian~Hao,~\IEEEmembership{Student Member,~IEEE,}
        Ashwin~Lele,~\IEEEmembership{Member,~IEEE,}
        
        Yan~Fang,~\IEEEmembership{Member,~IEEE,}
        Arijit~Raychowdhury,~\IEEEmembership{Fellow,~IEEE,}
        and~Azadeh~Ansari,~\IEEEmembership{Member,~IEEE}
\thanks{This work was supported in part by Semiconductor Research Corporation's (SRC) COCOSYS Center under Grant SRC JUMP 2.0. \textit{(Corresponding author: Azadeh Ansari.)}}
\thanks{Z. Hao, A. S. Lele, A. Raychowdhury, and A. Ansari are with the Department of Electrical and Computer Engineering, Georgia Institute of Technology, Atlanta, GA, 30332 USA (e-mail: \{zhao38, alele9\}@gatech.edu; \{arijit.raychowdhury, azadeh.ansari\}@ece.gatech.edu.}
\thanks{Y. Fang is with Department of Electrical and Computer Engineering, Kennesaw State University, Marietta, GA, USA (email: yfang9@kennesaw.edu).}
}

\maketitle

\begin{abstract}
Robotic autonomy at centimeter scale requires compact and miniaturization-friendly actuation integrated with sensing and neural network processing assembly within a tiny form factor. Applications of such systems have witnessed significant advancements in recent years in fields such as healthcare, manufacturing, and post-disaster rescue. The system design at this scale puts stringent constraints on power consumption for both the sensory front-end and actuation back-end and the weight of the electronic assembly for robust operation. In this paper, we introduce FAVbot, the first autonomous mobile micro-robotic system integrated with a novel actuation mechanism and convolutional neural network (CNN) based computer vision - all integrated within a compact 3-cm form factor. The novel actuation mechanism utilizes mechanical resonance phenomenon to achieve frequency-controlled steering with a single piezoelectric actuator. Experimental results demonstrate the effectiveness of FAVbot's frequency-controlled actuation, which offers a diverse selection of resonance modes with different motion characteristics. The actuation system is complemented with the vision front-end where a camera along with a microcontroller supports object detection for closed-loop control and autonomous target tracking. This enables adaptive navigation in dynamic environments. This work contributes to the evolving landscape of neural network-enabled micro-robotic systems showing the smallest autonomous robot built using controllable multi-directional single-actuator mechanism.\end{abstract}


\IEEEpeerreviewmaketitle

\section{Introduction}

\begin{figure}[t!]
    \centering
  \subfloat[\label{fig:intro_plot_group_a}]{%
       \includegraphics[width=1.0\linewidth]{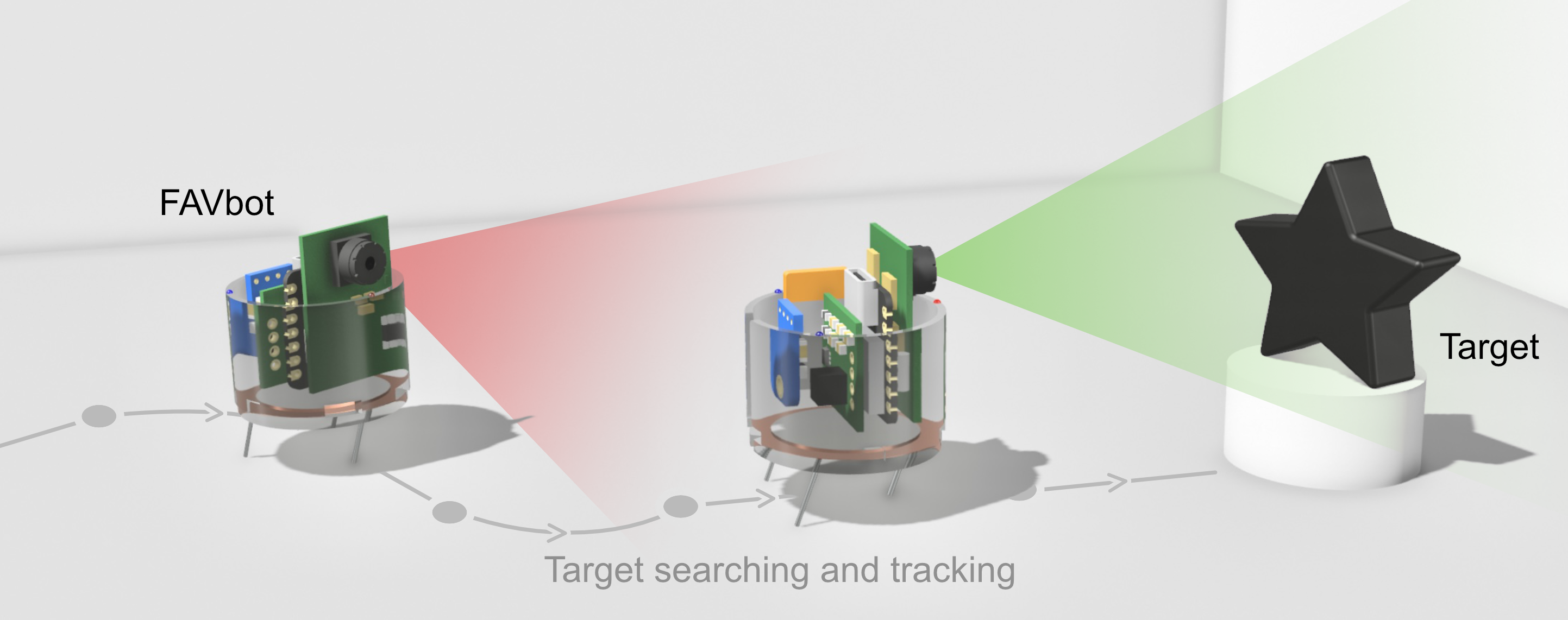}}
        \\
    \subfloat[\label{fig:intro_plot_group_b}]{%
       \includegraphics[height=3.7cm]{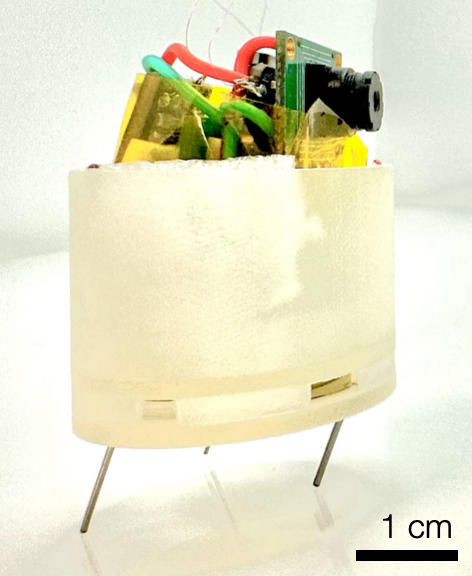}}
    \subfloat[\label{fig:intro_plot_group_c}]{%
       \includegraphics[height=3.7cm]{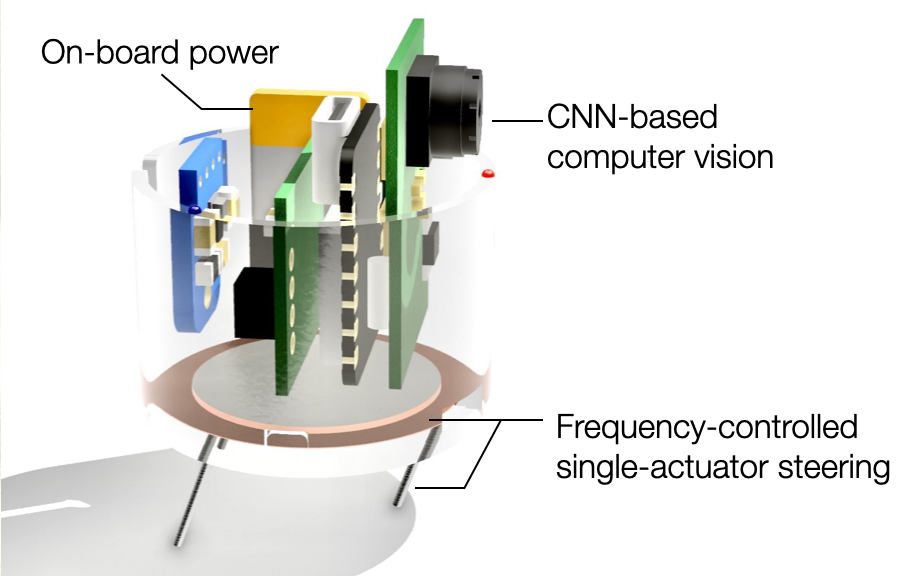}}
       \hfill
        \\

  \caption{FAVbot: 3-cm miniaturized robot. (a) Conceptual rendering of robot autonomous target tracking operation. (b) Image of the assembled robot. The diameter of the robot is 3 cm. (c) Sub-modules of the robot, including on-board power, CNN-based computer vision, and frequency-controlled single-actuator steering.}
  \label{fig:intro_plot_group} 

\end{figure}

\IEEEPARstart{I}{n} the ever-evolving landscape of robotics, the demand for miniaturized robots has intensified, driven by the need for versatile, agile, and adaptable systems in various applications ranging from healthcare \cite{lee2018capsule}, manufacturing \cite{cappelleri2014towards, limeira2019wsbot} to post-disaster search and rescue \cite{misaki2011development}, as well as environmental exploration and monitoring \cite{schaler2022swim}.
However, the miniaturization of robots poses a unique set of challenges, notably the integration of enabling functionalities: sensing, actuation / motion control, computation, communication, and power, limiting the scope of applications for small-scale robotic systems \cite{Caprari_Estier_Siegwart_2001}. This paper addresses the actuation and control challenge by presenting an autonomous mobile micro-robot, FAVbot (Frequency Actuated with Vision robot), in 3-cm size (Fig.~\ref{fig:intro_plot_group}) combining the end-to-end sensing to processing assembly. We introduce a novel frequency-controlled single-actuator steering mechanism combined with a convolutional neural network (CNN) vision program, \new{as inspired by frequency-modulated motions in many insect species~\cite{gau2021rapid}.}

\begin{table*}[]
\centering
\caption{\new{Power autonomous MMRs comparison}}
\label{table:power-autonomous-MMRs}
\resizebox{\textwidth}{!}{%
\begin{tabular}{l||rrr|ccl|rrr|cc}
\hline
Robot &
  \begin{tabular}[c]{@{}r@{}}Lateral size\\ (cm)\end{tabular} &
  \begin{tabular}[c]{@{}r@{}}Speed\\ (cm/s)\end{tabular} &
  \begin{tabular}[c]{@{}r@{}}Speed\\ (BL/s)\end{tabular} &
  \begin{tabular}[c]{@{}c@{}}Control\\ DoF\end{tabular} &
  \begin{tabular}[c]{@{}c@{}}Actuator\\ Count\end{tabular} &
  Actuator &
  \begin{tabular}[c]{@{}r@{}}Runtime\\ (min)\end{tabular} &
  \begin{tabular}[c]{@{}r@{}}Power\\ (W)\end{tabular} &
  \begin{tabular}[c]{@{}r@{}}Weight\\ (g)\end{tabular} &
  Camera &
  CNN \\ \hline
FAVbot (This work) &
  \cellcolor[HTML]{DDFFC8}3.0 &
  \cellcolor[HTML]{DDFFC8}6.9 &
  \cellcolor[HTML]{DDFFC8}2.3 &
  2 &
  \cellcolor[HTML]{DDFFC8}1 &
  Piezoelectric buzzer &
  \cellcolor[HTML]{DDFFC8}15 &
  \cellcolor[HTML]{DDFFC8}$<$ 1.2 &
  21.7 &
  \cellcolor[HTML]{DDFFC8}Y &
  \cellcolor[HTML]{DDFFC8}Y \\
Alice \cite{Caprari_Estier_Siegwart_2001} &
  \cellcolor[HTML]{DDFFC8}2.1 &
  4.0 &
  2.0 &
  2 &
  2 &
  Electrical motors &
  \cellcolor[HTML]{DDFFC8}600 &
  \cellcolor[HTML]{DDFFC8}0.010 &
  5.0
   &
   \\
Insect-scale robot \cite{iyer2020wireless} &
  \cellcolor[HTML]{DDFFC8}1.6 &
  3.5 &
  2.2 &
  2 &
  2 &
  Vibration motors &
  \cellcolor[HTML]{DDFFC8}$>$ 60 &
  \cellcolor[HTML]{DDFFC8}0.033 &
  2.8 &
  \cellcolor[HTML]{DDFFC8}Y &
   \\
GRITSBot \cite{pickem2015gritsbot} &
  \cellcolor[HTML]{DDFFC8}3.0 &
  \cellcolor[HTML]{DDFFC8}25.0 &
  \cellcolor[HTML]{DDFFC8}8.3 &
  2 &
  2 &
  Electrical motors &
  \cellcolor[HTML]{DDFFC8}63 &
  &
  &
   &
   \\
RoACH \cite{hoover2008roach} &
  \cellcolor[HTML]{DDFFC8}3.0 &
  3.0 &
  1.0 &
  2 &
  2 &
  Shape memory alloy &
  9 &
  &
  2.4 &
   &
   \\
RoboAnt \cite{mclurkin1996using} &
  $^* >$ 3.2 &
  \cellcolor[HTML]{DDFFC8}15.2 &
   &
  2 &
  2 &
  Electrical motors &
  \cellcolor[HTML]{DDFFC8}20 &
  \cellcolor[HTML]{DDFFC8}0.552 &
  36.9 &
   &
   \\
Kilobot \cite{rubenstein2012kilobot} &
  3.3 &
  ~ 1.0 &
  ~ 0.3 &
  2 &
  2 &
  Vibration motors &
  \cellcolor[HTML]{DDFFC8}180 &
  &
  &
   &
   \\
MARS \cite{fukuda1999group} &
  3.5 &
   &
   &
  2 &
  2 &
  Electrical motors &
   &
   &
   &
   &
   \\
HAMR \cite{Goldberg_Zufferey_Doshi_Helbling_Whittredge_Kovac_Wood_2018} &
  4.5 &
  \cellcolor[HTML]{DDFFC8}17.2 &
  \cellcolor[HTML]{DDFFC8}3.8 &
  1 &
  2 &
  Piezoelectric bimorph &
  4.5 &
  \cellcolor[HTML]{DDFFC8}0.5 &
  2.8 &
   &
   \\
Millibot \cite{navarro1999modularity} &
  6.3 &
   &
   &
  2 &
  2 &
  Electrical motors &
  \cellcolor[HTML]{DDFFC8}90 &
  &
  &
  \cellcolor[HTML]{DDFFC8}Y &
   \\
   \hline
\multicolumn{8}{l}{* Exact value is not reported. Green fill color indicates better or equal trait as compared to this work.}
\end{tabular}%
}
\end{table*}

Existing works on mobile micro-robots (MMR) exhibit a wide spectrum of size ranging from sub millimeter to tens of centimeters with a variety of actuation mechanisms \cite{hussein2023actuation}. 
MMRs with form factors smaller than 1 cm are normally not equipped with on-board power and either require tethers to a power supply \cite{yang2019bee, fuller2019four, de2018inverted}, or are powered externally using magnetic fields \cite{vogtmann201725, kim2020magnetically}, acoustic waves \cite{kaynak2017acoustic, kim20195, hao2022controlling}, light \cite{zeng2018light}, etc. The sensing and computation needed for closed-loop control, if any, are also performed externally. The size constraint on these robots puts stringent constraints on power and adds complexity in assembling a diverse set of electronic components within a tiny form factor. This has hindered the deployment of AI-infused intelligence on such size-constrained robots. For example, Wang \textit{et al.} controlled the path of a magnetically driven micro-robot by tracking robot location with microscopic imaging and adjusting the external magnetic field accordingly \cite{wang2022tetherless}. The applications of these externally actuated robots are limited by the range of the external power and the size of the workspace.
In contrast, current power-autonomous MMRs typically carry on-board batteries resulting in a larger size of a few centimeters \cite{Caprari_Estier_Siegwart_2001, iyer2020wireless, pickem2015gritsbot, hoover2008roach, mclurkin1996using, rubenstein2012kilobot, fukuda1999group,    Goldberg_Zufferey_Doshi_Helbling_Whittredge_Kovac_Wood_2018,  navarro1999modularity, berlinger2021implicit}. As shown in Table~\ref{table:power-autonomous-MMRs}, to achieve controllable motion, all centimeter-scale MMRs used 2 or more actuators for the degree of freedom (DoF) specific to their applications. 
For 2-D applications in ground robots, using differential driving of 2 actuators is still a popular choice among recently-published miniaturized micro-robots \cite{Caprari_Estier_Siegwart_2001, iyer2020wireless, pickem2015gritsbot, mclurkin1996using, rubenstein2012kilobot, fukuda1999group, navarro1999modularity,      notomista2019study}. Due to the development of small COTS (commercial off-the-shelf) components, using electrical motors remains a viable solution in MMRs at a few-centimeter scale. However, both the need of multiple actuators and the intricate design of motors would prevent further miniaturization to the sub-centimeter or smaller down to sub-millimeter sizes.

One way to address the need for multiple components for actuation is using mechanical resonances where multiple resonance modes, distinct in the frequency domain, can be excited within the same structure.
\new{This concept takes inspirations from the nature, where many insect species use frequency modulation at near resonance to optimize and control motion through spontaneous adjustments of the frequency of their neural activities~\cite{gau2021rapid, pons2022distinct}.
In the inanimate world, frequency modulation is also applied in micro-electro-mechanical system (MEMS) resonators~\cite{uranga2015cmos} used for amplified sensitivity sensing and signal filtering applications as well as in piezoelectric motors~\cite{spanner2016piezoelectric} for precise motion control. However, utilizing the resonance frequency for motion control has not been demonstrated in power-autonomous MMRs.}

Such dependency of frequency is also observed in magnetically- or acoustically-driven sub-millimeter micro-robots, where the velocity depends on the frequency of the alternating actuation fields~\cite{vogtmann201725, kaynak2017acoustic,  supik2023magnetic}. By taking advantage of different resonance modes, a single actuator can induce different motion patterns (steering, speed control, etc.) with frequency control~\cite{dharmawan2017steerable}. \new{In these implementations, the robots are not power autonomous and operates in an open-loop fashion, resulting in limited applications. In comparison, our FAVbot is the first frequency-controlled MMR integrated with on-board power and driver for fully untethered actuation as well as the capability to operate autonomously with the closed-loop vision feedback.}

To excite multiple vibration modes for steering, we designed a micro-bristle-robot as shown in Fig.~\ref{fig:intro_plot_group}, where a piezoelectric actuator is used for generating vibration across a wide range of frequency from 1 to 62~kHz. In addition, the bristles of the robot are designed asymmetrically to be in and out of resonance at different frequencies based on their geometries. Bristle-robot's locomotion is achieved with stick-slip cycles of the bristles under vibration. Such behavior is modeled in previous work \cite{becker2014mechanics, cicconofri2015motility}. In confined spaces, bristle-robots have been demonstrated in applications including structural health inspection as in pipes \cite{gmiterko2002pipe, fath2024marsbot}, earthquake rescue \cite{wang2002simulation}, and healthcare as in colonoscopy  \cite{zhang2023development}. In addition, the simplicity of bristle-robot actuation makes it suitable for rapid prototyping. Kilobot by Rubenstein \textit{et al.} is a scalable and cost-effective example of hundreds of micro bristle-robots working collaboratively \cite{rubenstein2012kilobot}. Iyer \textit{et al.} used an insect-scale bristle-robot to demonstrate the functionality of their custom camera system \cite{iyer2020wireless}. However, none of these bristle robots used resonance-based actuation to simplify their actuation. Thus, they all require multiple actuators for differential steering.

\new{The FAVbot is developed from our previous work}, where we have demonstrated a tethered bristle robot with a piezoelectric actuator moving in different directions depending on the input frequency \cite{hao2020maneuver}. However, random variations in the heading direction is observed with the open-loop actuation in existing work \cite{supik2023magnetic}. This is due to the inherent stochasticity of vibration actuation, as well as the sensitivity of resonance to the environment and the change of internal structure of the robot (e.g., a shift of the center of mass) \cite{becker2014mechanics}. To address these nonidealities for reliable motion control, our FAVbot is integrated with a vision system for closed-loop operation (Fig.~\ref{fig:intro_plot_group_a}). The integration of vision not only corrects the motion nonideality in real-time, but also enables FAVbot to operate fully autonomously as later shown in the target tracking application, \new{making the frequency-controlled actuation a more robust mechanism.} This work presents the smallest vision-based autonomous robotic system to the best of our knowledge.

The main contributions of this paper are as follows:

\begin{itemize}
    \item First, the robot achieves actuation and steering through a novel single-actuator frequency-controlled mechanism. The utilization of mechanical resonance minimizes the complexity of actuation systems, allowing for further miniaturization.
    \item Second, on-board convolutional neural network (CNN) based computer vision program is integrated, enabling the robot to autonomously track and respond to a target in its environment. This capability is also a complement to the actuation mechanism for closed-loop control.
    \item Finally, the robot achieves a compact integration of all components in a 3-cm footprint, opening avenues for deployment in constrained spaces where traditional robots cannot operate effectively.
\end{itemize}

\section{Robot Design}

\begin{figure}[t!]
    \centering
    \subfloat[\label{fig:electronic_package_a}]{%
       \includegraphics[width=1.0\linewidth]{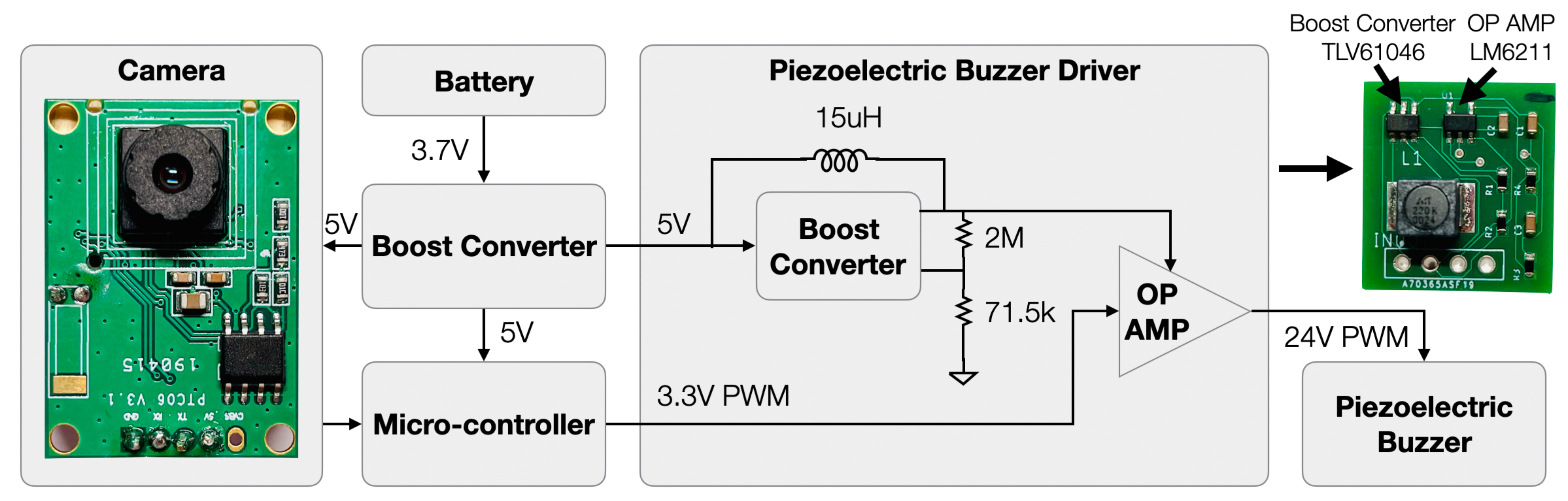}}
        \\
    \subfloat[\label{fig:electronic_package_b}]{%
       \includegraphics[width=1.0\linewidth]{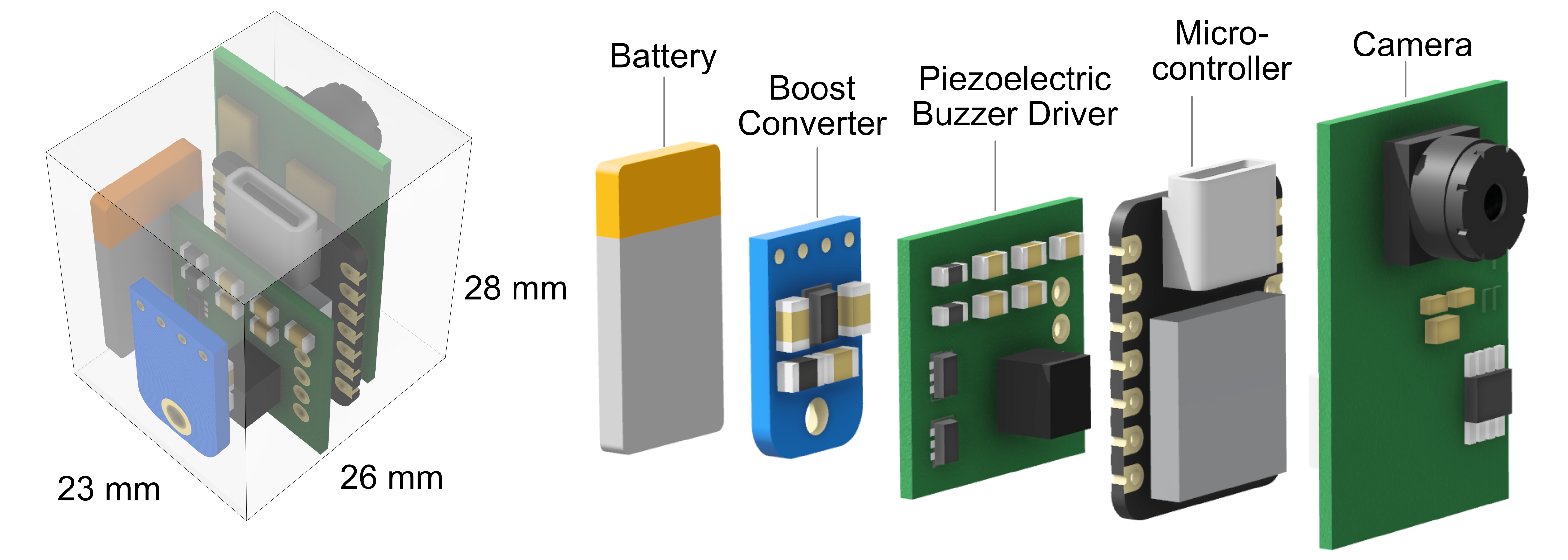}}
       \caption{(a) Circuit diagram. (b) Circuit components. COTS components has been used except the custom driver. The circuit component assembly (without the piezoelectric buzzer) occupies less than 23 $\times$ 26 $\times$ 28 mm$^3$ volume.}
       \label{fig:electronic_package}
\end{figure}

\begin{figure}[t!]
    \centering
    \includegraphics[width=0.8\linewidth]{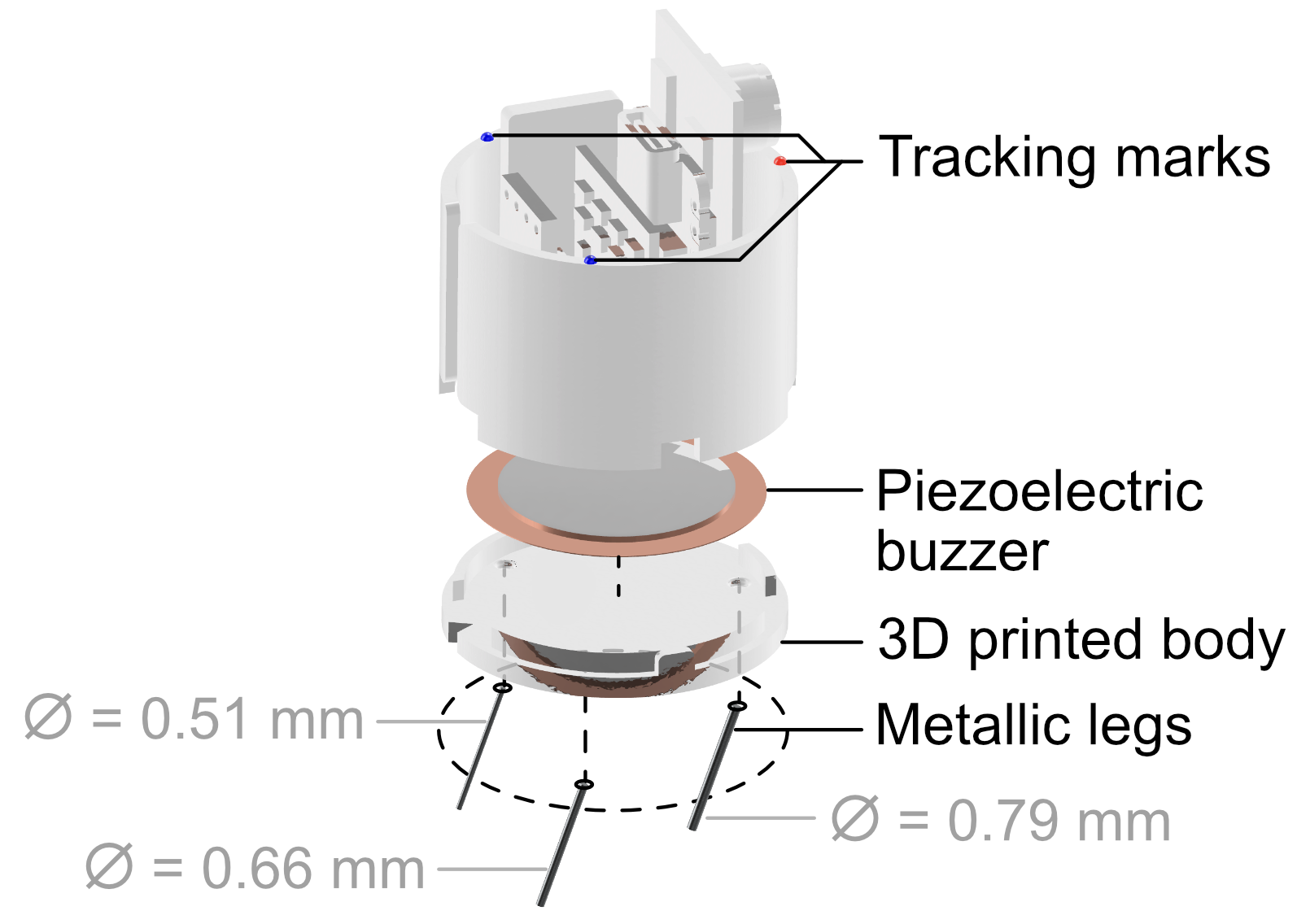}
       \caption{Mechanical design of FAVbot.}
       \label{fig:mechanical_a}
\end{figure}

The robot consists of an electronic system for capturing images and driving the piezoelectric buzzer for actuation (Fig.~\ref{fig:electronic_package}), and a mechanical system for locomotion under the buzzer vibration to enable frequency-controlled steering (Fig.~\ref{fig:mechanical_a}). The robot is in cylindrical shape with a diameter of 3~cm and a height (including the bristles) of 4.5~cm.

The electrical components (excluding the piezoelectric buzzer) are integrated in a volume of 23 mm $\times$ 26 mm $\times$ 28 mm. This work uses all COTS components except for the custom driver board, as detailed below: 

\begin{itemize}
    \item Sensor: A JPEG camera (Putal PTC06) which can snap images with a 60$^\circ$ field of view (FoV) and send them to the micro-controller over TTL serial link. The camera images are inputs to a convolutional neural network computer vision model, which enables closed-loop control and corrects actuation imperfections during autonomous operations.
    \item Processor: A Bluetooth-enabled micro-controller (Seeeduino XIAO nRF52840) for processing the computer vision model and generating frequency signals for actuation.
    \item Power Source: A 3.7 V, 40 mAh lithium battery (Sparkfun PRT-13852) together with a boost converter (Adafruit 4654) to power the components, which supports 15 minutes of untethered actuation.
    \item Actuator: A custom driver board that amplifies the signal from the micro-controller to 24 V to drive the piezoelectric buzzer (CUI Devices CPT-2746-L100), which generate vibration at varying frequencies for actuation (Fig.~\ref{fig:electronic_package}, inset).
    \item Platform: A piezoelectric buzzer (CUI Devices CPT-2746-L100) to generate on-board vibration at different frequencies.
\end{itemize}

\begin{figure}[t!]
    \centering
    \includegraphics[width=\linewidth]{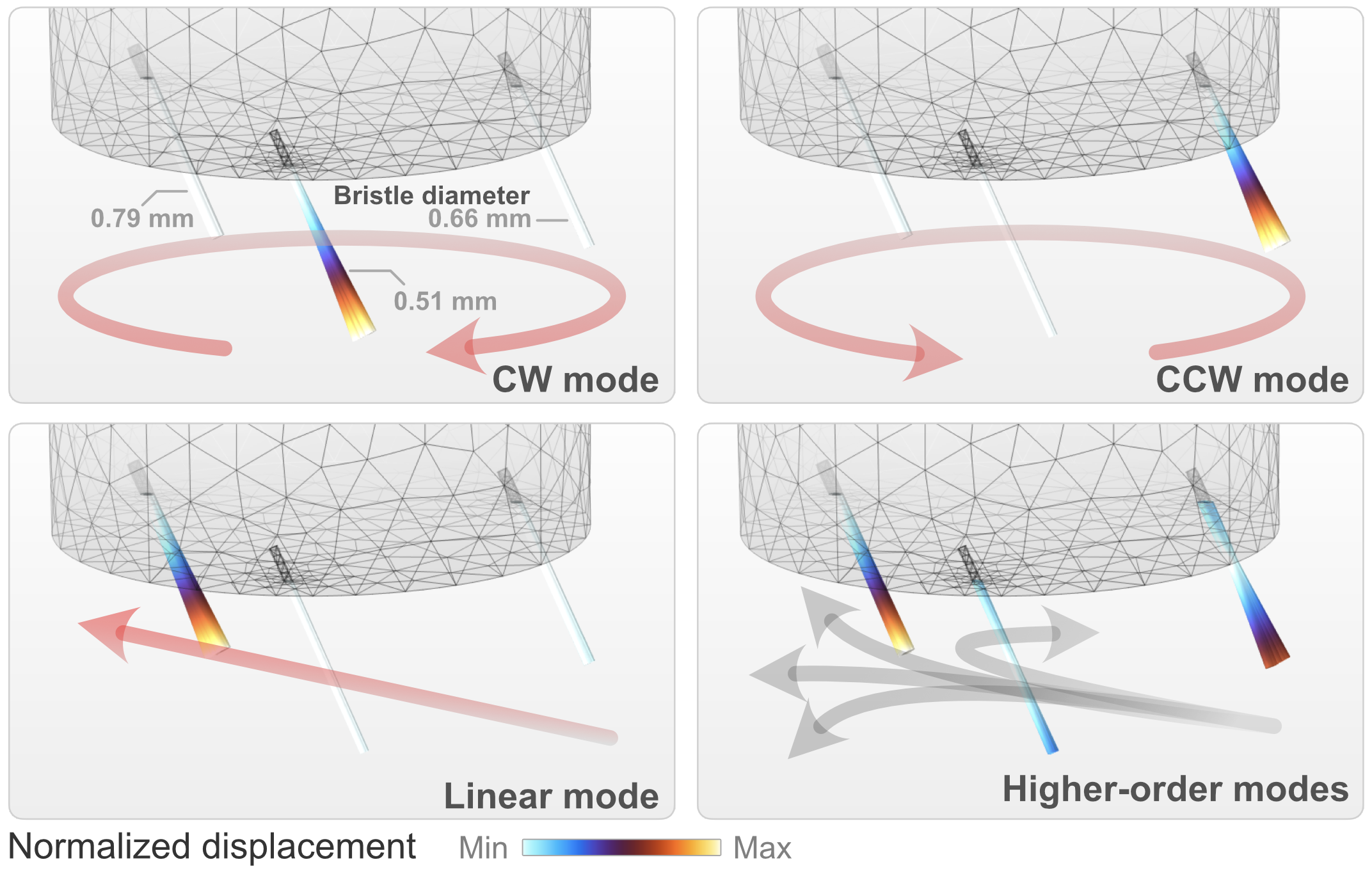}
       \caption{Resonance mode shapes from finite elements analysis.}
       \label{fig:comsol}
\end{figure}

\begin{table}[ht!]
\centering
\caption{Weight and power consumption of each robot component}
\label{table:component_weight_and_power}
\begin{tabular}{lrr}
\hline
Component            & Weight (g) & Power (W) \\ \hline
JPEG camera          &      2.0      &       0.375                 \\
Micro-controller     &      2.1      &          0.660              \\
Custom driver board  &      1.6      &      0.007$^a$      \\
Piezoelectric buzzer &      2.0      &      At 1~kHz: 0.004$^b$     \\
                      &        &         At 100~kHz: 0.020$^b$  \\
Boost converter      &       1.4        &      0.140$^c$           \\
Battery              &     5.0       &    N/A                    \\
3D printed body      &     6.6    &    N/A                    \\
Metallic bristles        &     0.1    &    N/A                    \\ 
Misc. (wiring, tape, etc.)        &    2.5     &    N/A                    \\ \hline
Total & 21.7 & 1.202
\\ \hline
\multicolumn{3}{l}{
    \begin{tabular}[c]{@{}l@{}}
    \footnotesize
    $^a$Estimated based on power consumption of individual components.\\ 
    $^b$Estimated by frequency multiplied by each cycle's stored energy\\in the piezoelectric material, $P = f\times \frac{1}{2} \times C \times V^2$.\\ 
    $^c$Estimated by total power multiplied by component efficiency.
    \end{tabular}
}
\end{tabular}
\end{table}

\new{The electronics, located in the top holder, is distanced from the actuation system to minimize the heating impact.}

The mechanical design (Fig.~\ref{fig:mechanical_a}) consists of two 3D printed parts, one for holding the on-board electronics, and the other for assembling the piezoelectric actuator and metallic bristles. Three stainless steel (SAE 304) wires with length of 12~mm and different diameters of 0.51, 0.66, and 0.79~mm are attached to the main body \new{at a tilt angle of 20$^{\circ}$. As opposed to conventional bristle-robots with much more bristles, the 3-bristle design ensures that all bristles are in contact with the surface, offering better consistency and controllability of the resonance. The use of stainless steel bristles provide better structural rigidity to carry the payload of electronics as well as higher mechanical quality factor for more pronounced resonance behaviors. The bristles' length, diameters, and the tilt angle affects the resonance of the system. The final configuration is determined based on the finite element analysis (FEA) of the mechanical system as shown in Fig.~\ref{fig:comsol}, where distinct resonance modes are observed at well-distanced frequencies over the actuation range.}

Under vibration the bristles go through stick-slip cycles to generate locomotion~\cite{becker2014mechanics,cicconofri2015motility}. The different bristle diameters provide asymmetrical spring constants which lead to the rich resonance modes and different motion characteristics across frequency~\cite{hao2020maneuver}. The inclination of the bristles provides forward directionality, but the response among bristles could be different under specific frequencies. 

When actuated at the resonance frequency of one bristle (e.g. left or right), the displacement of the particular bristle is amplified. Analogous to differential drive, the difference in vibration amplitudes between the left and right bristles cause the robot to steer in the left or right direction.

The frequency dependent behavior has been modeled and observed experimentally in our previous work~\cite{hao2020maneuver} as well as in magnetically-driven micro-bristle-robot by Supik \textit{et al.}~\cite{supik2023magnetic}. In addition, it has been observed that resonance could cause both forward and backward motion~\cite{kim2020forward} as well as motion orthogonal to the inclination~\cite{hao2020maneuver}. \new{However, these models assume that all bristles are either identical and operate synchronously or grouped in sets among which the resonance is uncoupled. In addition, these models are agnostic to the robot's geometry, which falls short in predicting the motion patterns of our 3-bristle design with varying bristle geometries and their couplings. Instead, in this work we utilize finite element analysis conducted with the COMSOL Multiphysics software. The simulation models the robot as a whole structure to determine the resonance mode shapes of the bristles. As shown in Fig.~\ref{fig:comsol}, first-order modes are observed at distinct frequencies for individual bristles. These modes are related to clockwise (CW), counterclockwise (CCW), and linear motions. In addition, higher-order modes in higher frequency ranges are observed. The combined contribution of all three bristles at these frequencies can enable rich motion characteristics. This frequency-controlled motion is characterized in the later section.}

The weight and power breakdown of the robots is summarized in Table~\ref{table:component_weight_and_power}. Please note that the camera is only active during image capturing, thus the average power consumption during operation is lower than the listed value and depends on the duration of the vision pipeline as compered to actuation segments.

\section{Frequency-Controlled Motion}\label{sec:motion}

As mentioned earlier, the robot has an asymmetrical design to induce a variety of resonance modes among the different bristle geometries. This frequency-dependent actuation is characterized by sweeping the actuation frequency from 1 to 100~kHz \new{with a step size of 1 kHz.}

\new{During testing, all the components are added to ensure consistency in payload and component placement. To support the long operation duration required for characterization, the power is supplied externally by 42 AWG magnet wires with minimal tension and do not impedes motion.} The frequency command is sent wirelessly via Bluetooth. More details on the communication protocol is discussed in Section~\ref{sec:close-loop}. \new{The substrate which the robot operates on has a significant impact on the motion characteristics. In our experiment, the robot motion is characterized on a glass substrate, which is known to provide proper frictional characteristics for the stick-slip cycle~\cite{kim2020forward}. While the robot is capable to operate on other substrates such as Teflon and silicon, the speed and controllability are less optimal as compared to those on glass.}

\begin{figure*}[t!]
    \centering
    \subfloat[\label{fig:motion_panel_a}]{%
       \includegraphics[width=0.95\linewidth]{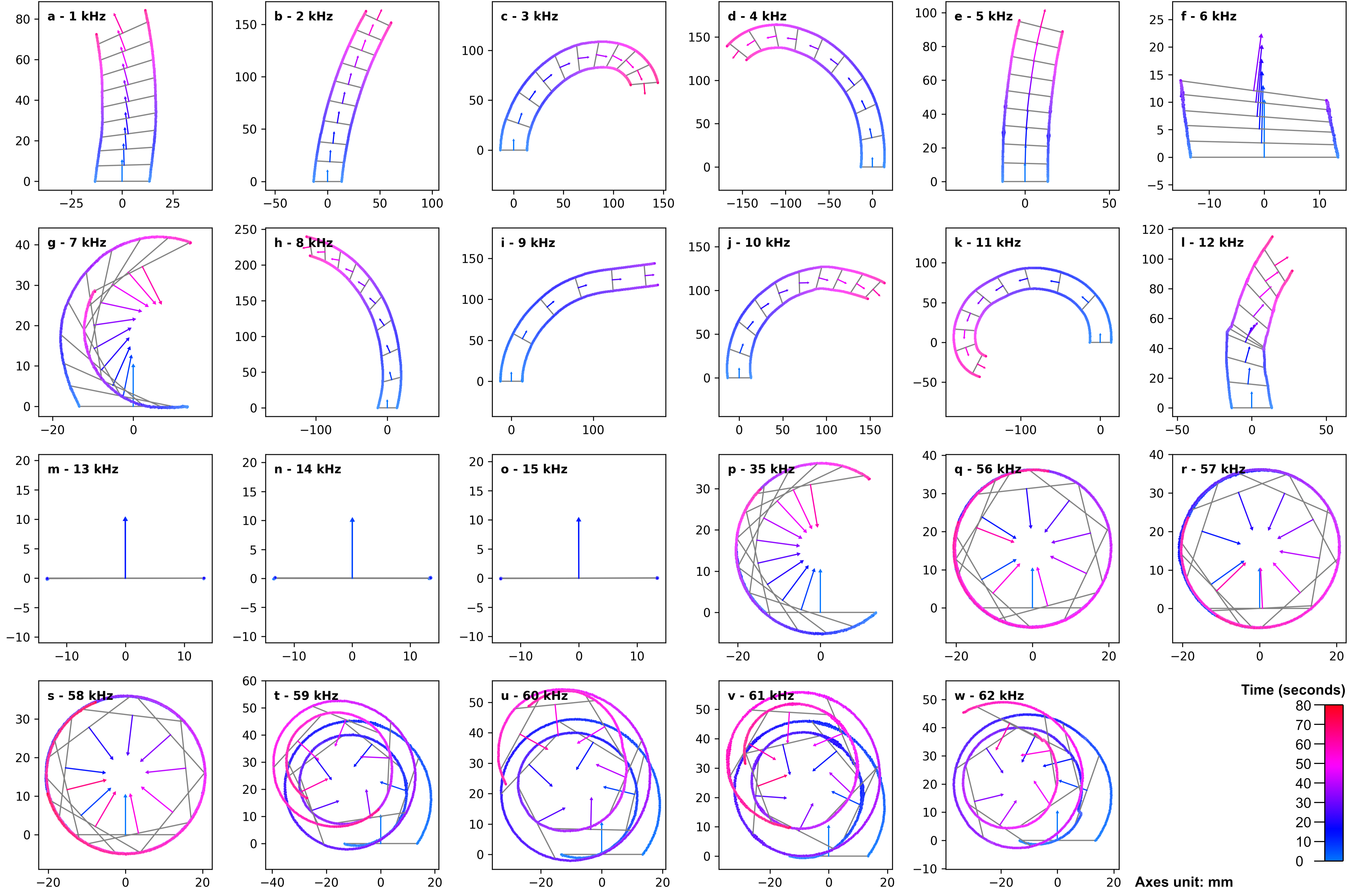}}
        \\
    \subfloat[\label{fig:motion_panel_b}]{%
       \includegraphics[width=0.95\linewidth]{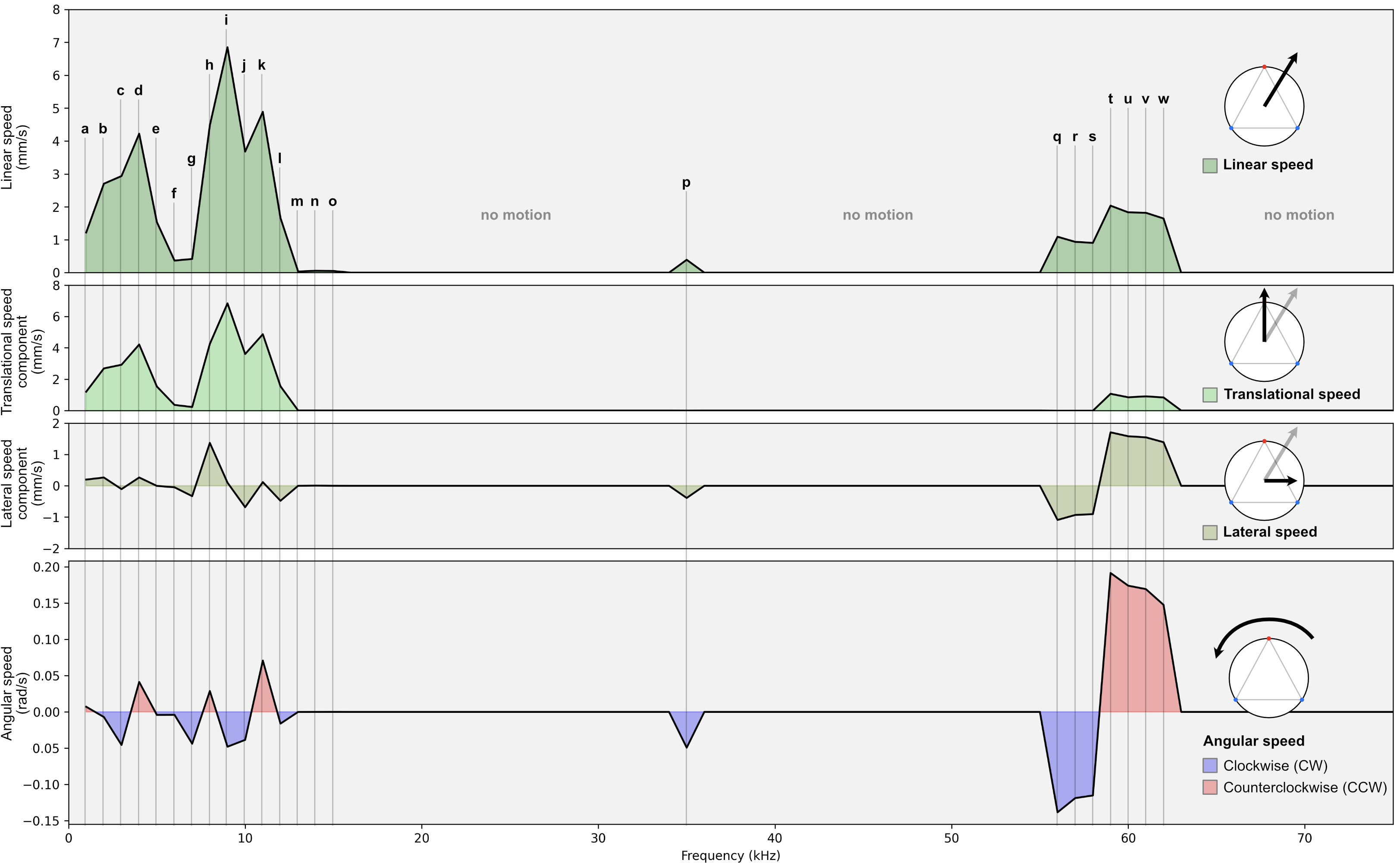}}

       \caption{Frequency-controlled steering. (a) A selection of representative motion patterns of FAVbot under various frequencies between 1 - 75~kHz. Arrows plot the instantaneous orientations of the robot. Color represents the elapsed time. (b) Extracted average values of the linear, transnational, lateral, and angular speed of the robot under different frequencies. Labels a - w are consistent in both plots. Videos of some modes can be found in the multimedia attachment.}
       \label{fig:motion_panel}
\end{figure*}

Fig.~\ref{fig:motion_panel_a} summarizes a few representative modes of motion across the frequency spectrum. The two tracking marks at the rear of the robot (blue dots in Fig.~\ref{fig:mechanical_a}) are used to extract robot trajectories. The two marks are plotted in color gradient, which indicates time progression. Connection lines between the marks are added every 200 frames (or 6.67 seconds) for clarity, and a vector perpendicular to the connection line visualizes robot orientation / heading at the given moment.

The modes shown in Fig.~\ref{fig:motion_panel_a} exhibit distinct motion characteristics, providing multiple combinations of linear speed (including various translational and lateral compositions), angular speed, and radius of curvature to choose from. Qualitatively, the mode at 5~KHz is most suitable for forward motion with the best directionality due to the balance of rotational modes in opposite directions (at 4 and 7~kHz). In comparison, modes under 2~kHz and 8~kHz offer forward motion but coupled with slight steering in the opposite side, while 56-58~kHz and 59-62~kHz produce tight clockwise (CW) and counter-clockwise (CCW) motions with small radius of curvature. The existence of no-motion band starting at 13~kHz further emphasizes the underlying mechanical resonance phenomenon.

\begin{figure*}[h!]
    \centering
       \includegraphics[width=\linewidth]{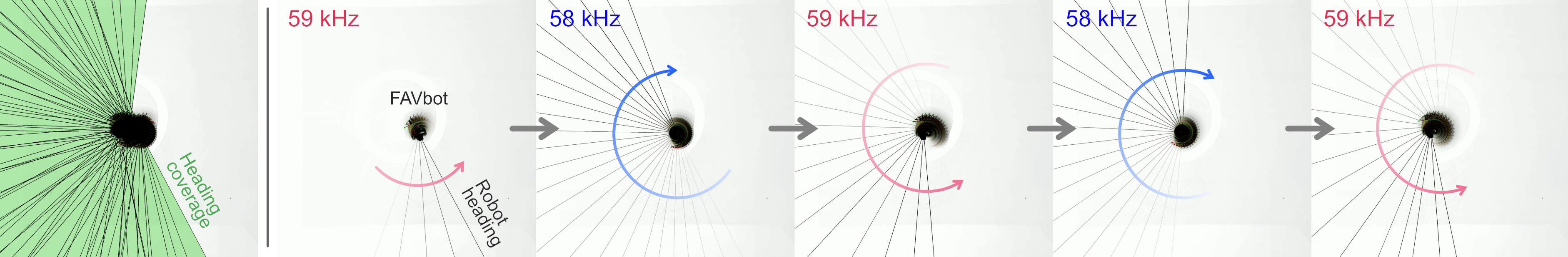}
       \caption{Alternating resonance modes at 58~kHz and 59~kHz to achieve CW and CCW motion, respectively, with minimal linear motion for surrounding scouting. First image shows the total angular coverage of the actuation sequence on the left. Experimental video can be found in the multimedia attachment.}
       \label{fig:58_59_alternating}
\end{figure*}

It is worth mentioning that discontinuities are observed in some trajectories (Fig.~\ref{fig:motion_panel_a}, case i, j, k). This implies that these modes are more sensitive to imperfections of the experimental setup such as minuscule obstacles or dips, variation in friction coefficients, and slight gravitational pull due to unlevelness of the substrate. In contrast, other modes in Fig.~\ref{fig:motion_panel_a} are robust against these nonidealities and exhibit arc-like trajectories. \new{This observation indicates the need for closed-loop control to correct the robot's orientation when these deviations happen which is enabled by FAVbot's vision system as detailed in later sections.}

Linear speed's absolute value $v$, translational component $v_t$ and lateral component $v_l$ to the orientation, as well as angular speed are extracted from the characterization experiment:

\begin{align}
    \dot{x_c}[\textrm{T}] &= (x_c[\textrm{T}+\delta \textrm{t}]-x_c[\textrm{T}]) / \delta t ,\\ 
    \dot{y_c}[\textrm{T}] &= (y_c[\textrm{T}+\delta \textrm{t}]-y_c[\textrm{T}]) / \delta t ,\\
    \dot{\theta}[\textrm{T}] &= \theta[\textrm{T}+\delta \textrm{t}]-\theta[\textrm{T}] / \delta t ,\\
    \va*{v}[\textrm{T}] &= (\dot{x_c}[\textrm{T}], \dot{y_c}[\textrm{T}]) ,\\
    v[\textrm{T}] &= \left|\left|\va*{v}[\textrm{T}]\right|\right| ,\\
    \va*{n}_t[\textrm{T}] &= (\cos(\theta[\textrm{T}]), \sin(\theta[\textrm{T}])) ,\\
    \va*{n}_l[\textrm{T}] &= (\cos(\theta[\textrm{T}]+\pi/2), \sin(\theta[\textrm{T}]+\pi/2)) ,\\
    v_t[\textrm{T}] &= \va*{v}[\textrm{T}] \cdot \va*{n}_t[\textrm{T}] / \left|\left|\va*{n}_t[\textrm{T}]\right|\right| ,\\
    v_l[\textrm{T}] &= \va*{v}[\textrm{T}] \cdot \va*{n}_l[\textrm{T}] / \left|\left|\va*{n}_l[\textrm{T}]\right|\right| ,
\end{align}

where $x_c$ and $y_c$ are global coordinates of the center point of the robot as computed from the tracking mark locations, $\theta$ is robot orientation, $\textrm{T}$ is time, $\delta \textrm{t}$ is time window, $\va*{v}$ is linear speed of the center point in global frame, $\va*{n}_t$ and $\va*{n}_l$ are unit vectors in the translational and lateral direction.

The quantitative motion characteristics are plotted against frequency in Fig.~\ref{fig:motion_panel_b}. Note that because $v$ is decomposed per frame before averaging in $v_t$ and $v_l$ computation, the vector sum of translational and lateral velocity is slightly different from the linear velocity. The data points are consistently labeled with those in Figure \ref{fig:motion_panel_a}. The results are summarized as follow:

\begin{enumerate}
    \item Maximum linear speed of 6.9~cm/s is achieved at (i)~9~kHz. As summarized in Table~\ref{table:power-autonomous-MMRs}, \new{both the absolute speed and the speed relative to the body length are better than many of the existing MMRs using multiple actuators.}
    \item Maximum angular speed of 0.19 rad/s (or 11.0 deg/s) is achieved at (t)~59~kHz, allowing for fast turning.
    \item It is evident that the motion is induced by resonance, as there exist a few no-motion bands (13-34~kHz, 36-55~kHz, and 63-100~kHz) between these modes, matching off-resonance behavior. (m-o)~13-15~kHz are plotted for reference while results for other frequencies are omitted.
    \item In the 1-12~kHz frequency band, both left and right resonances are competing against each other, causing alternating CW and CCW bands. Though the current frequency resolution is limited to be 1~kHz, theoretically straight motion could be achieved by balancing the CW and CCW motion, as demonstrated in (e)~5~kHz and (f)~6~kHz.
    \item Lateral drift motion where the major speed composition is tangential to the robot orientation is observed in (p-w)~35-62~kHz, such higher-order resonance has been reported in~\cite{hao2020maneuver}. 
    \item The motion in (p-w)~35-62~kHz is also contributed by the reverse in direction of one side of the robot, matching result and model in~\cite{kim2020forward}. When this happens, one side of the robot will move forward and the other side will move backward causing the robot to turn with a small radius of curvature.
\end{enumerate}

The capability to use higher-order resonance modes (e.g., under 58~kHz and 59~kHz) to scout the environment is demonstrated in Fig.~\ref{fig:58_59_alternating}, where the two frequencies are alternated. Images captured at different moments are overlaid in the first figure, where the coverage of robot heading is highlighted in green. The progression under each actuation segment is shown to the right. As shown, the robot stays relatively still in (x,y) position while changing its orientation. These modes are especially useful in search of targets during operation. 

The rich frequency-dependent motion pattern allows the robot to maneuver flexibly in complex terrines. Multiple of the resonance modes characterized in this section are used later to drive the robot in an autonomous object tracking application, demonstrating the practicality of the resonance-based motion control achieved with a single actuator.

\section{Computer Vision}\label{sec:vision}

\begin{figure}[t!]
    \centering
    \subfloat[\label{fig:CNN}]{%
       \includegraphics[width=\linewidth]{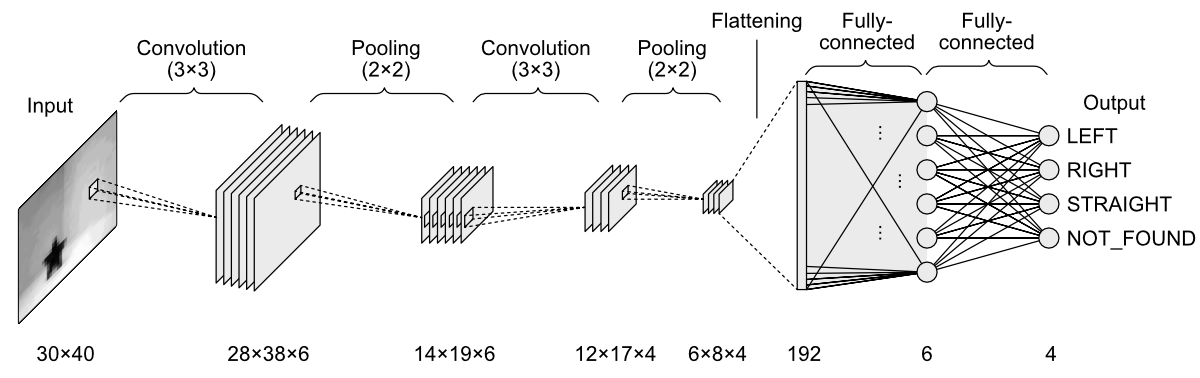}} \\
    \subfloat[\label{fig:CNN_training}]{%
       \includegraphics[width=0.88\linewidth]{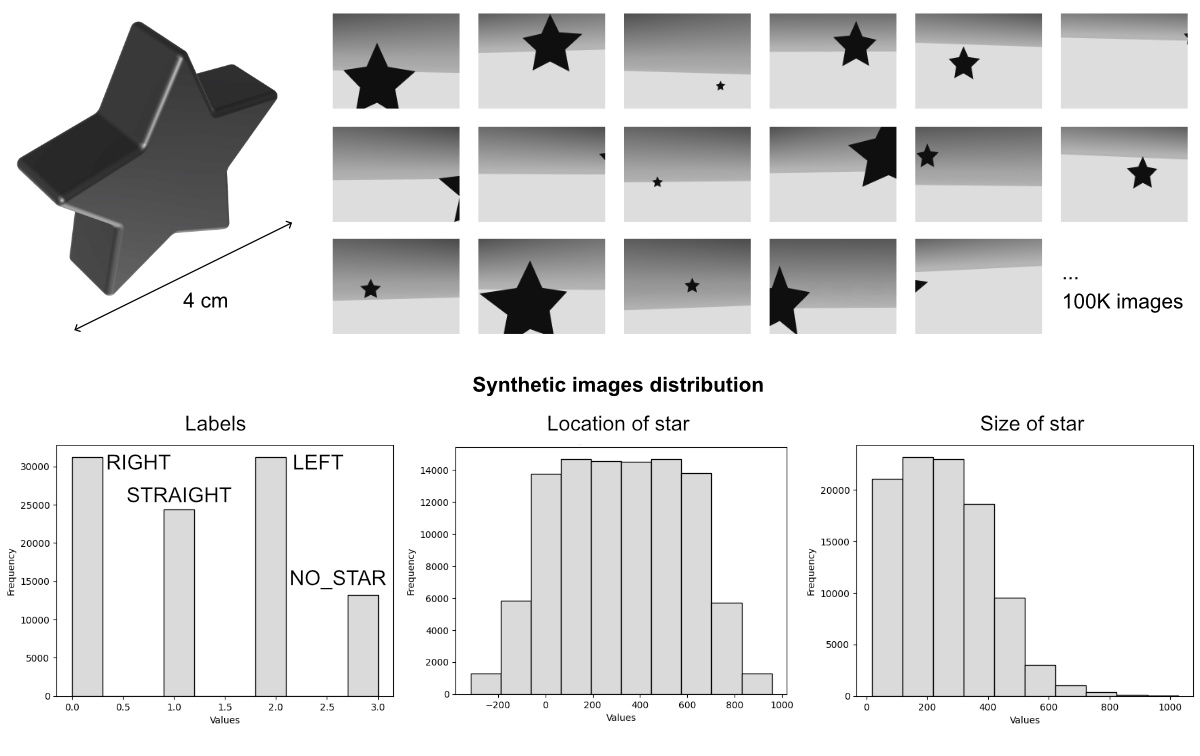}} \\
    \subfloat[\label{fig:cv_test}]{%
       \includegraphics[width=0.85\linewidth]{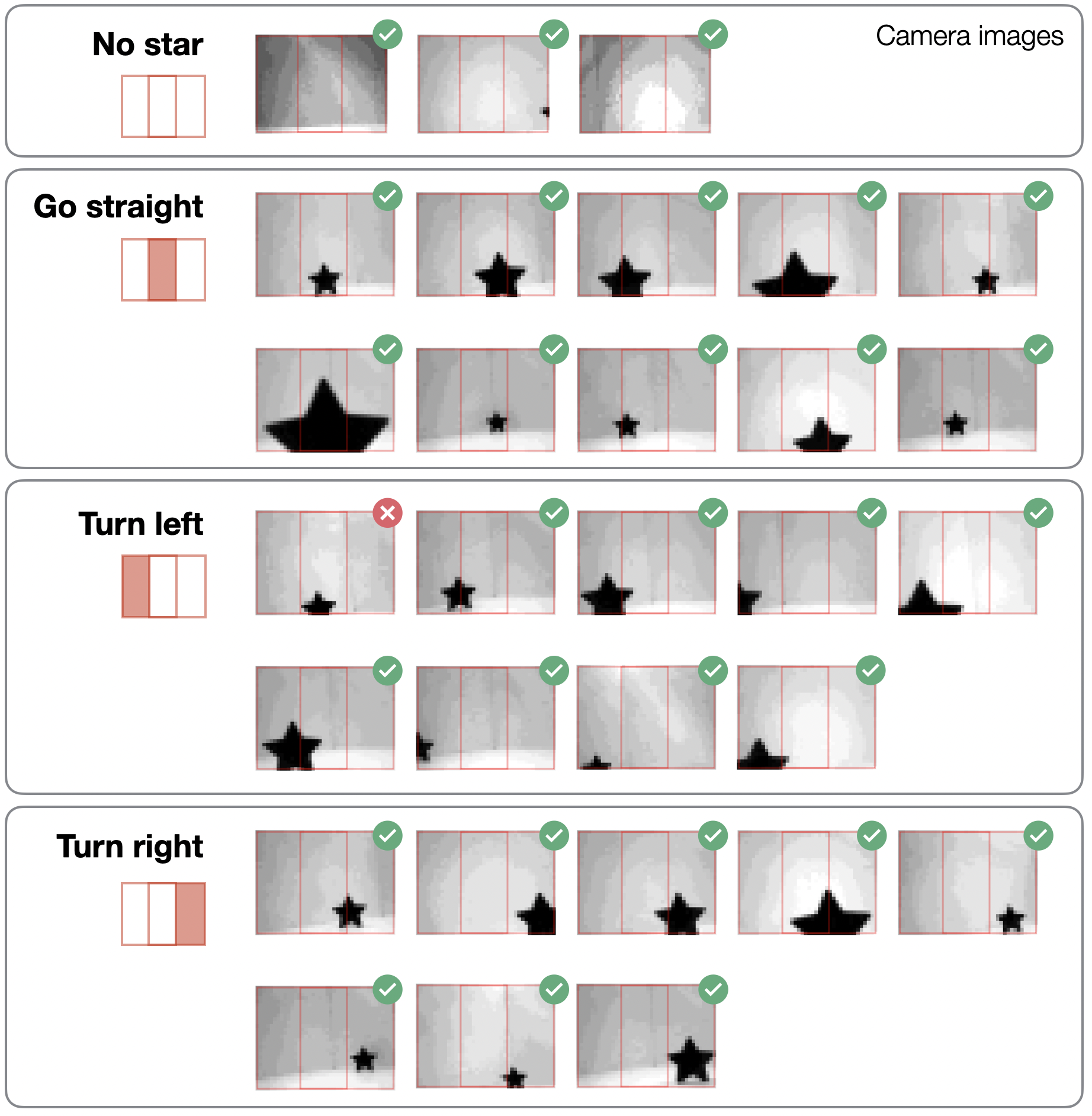}}
       
       \caption{(a) Convolutional neural network for object detection. (b) Computer vision tracking target and synthetic images for CNN training. Data distributions are plotted below. (c) Computer vision testing results on robot showing good accuracy.}
       \label{fig:CNN_group}
\end{figure}

In recent years, the integration of vision systems has emerged as a transformative addition to miniaturized robots. On top of imagery acquisition, the images could be used for autonomous operation. Our robot's control loop runs a pre-trained convolutional neural network (CNN) based computer vision program which enables it to autonomously search and track target objects with good accuracy. This closed-loop feedback corrects imperfections of the actuation mechanism and allows the robot to operate in dynamic environments. 

Structure of the neural network is shown in Fig.~\ref{fig:CNN}. The model is designed based on \new{the LeNet architecture capable of small-scale classifications to support real-time autonomy~\cite{lecun1998gradient}.} The model is implemented using TensorFlow and Keras libraries. The initial Conv2D layer employs six filters of size (3, 3) with ReLU activation, followed by a MaxPooling2D layer (2, 2) for spatial downsampling. A second Conv2D layer with four filters and subsequent MaxPooling2D layer further extract and refine features. The Flatten layer prepares the data for densely connected layers, comprising a 6-neuron Dense layer with ReLU activation and a final Dense layer with softmax activation, producing an output dimension equal to the number of actuation modes.

For demonstration, a star shaped object, as shown in Fig.~\ref{fig:CNN_training}, serves as the target for the robot to track. To train the neural network, 100k synthetic images are programmatically generated (Fig.~\ref{fig:CNN_training}), where a star is randomly placed, scaled, and skewed to represent 3D scenes as perceived by the camera. Each image is later down-sampled to 30 pixels by 40 pixels, matching the CNN input size. Each image is labeled according to the location of star within the image frame. Specifically, this positioning is categorized into four distinct zones: the left third, middle third, right third, and the area outside the frame. These zones are numerically represented as integers from 0 to 3, respectively (Fig.~\ref{fig:CNN_training}). \new{There is an inherent trade-off between image acquisition speed and the number of connections. To prioritize miniaturization, the serial camera is adopted. In experiments, the vision pipeline takes 3.59 $\pm$ 0.12 seconds on average to complete image capturing, down-sampling, and CNN inference. To avoid deterioration of the imaging quality, the vibration actuation is paused while the vision pipeline is executing.}

\begin{figure*}[!t]
    \centering
       \includegraphics[width=\linewidth]{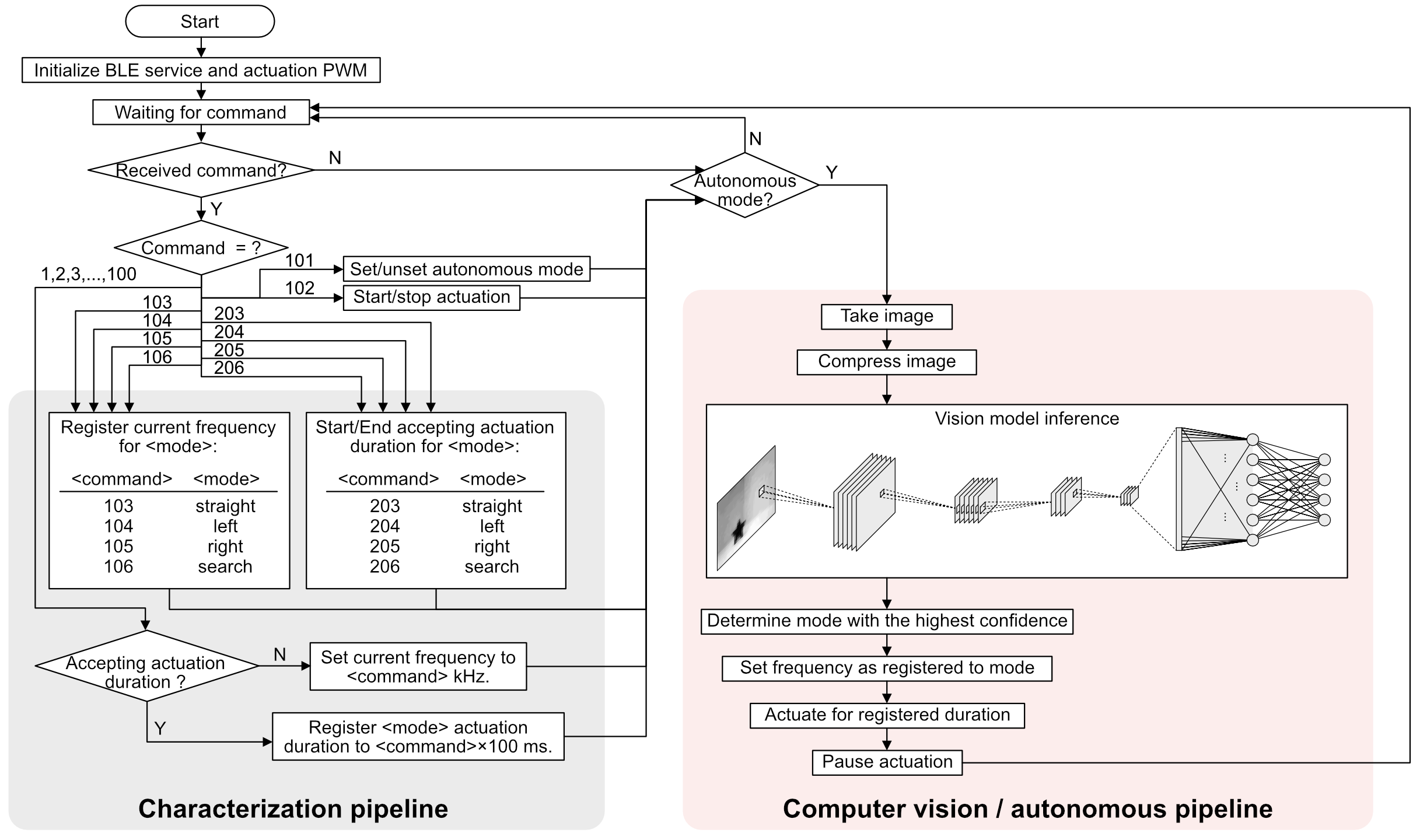}
       \caption{Robot software diagram showing the characterization pipeline and the computer vision / autonomous pipeline.}
       \label{fig:software_flow_diagram}
\end{figure*}

The model is compiled using the Adam optimizer and Sparse Categorical Crossentropy loss, trained for 20 epochs with a batch size of 64. \new{We use specific lighting conditions for training data and observe the validation accuracy of 96\%. However, previous LeNet works have been shown to carry out more complex tasks like selective classification with more cluttered environments and varying lighting conditions~\cite{liu2016combined}.} This pre-trained model is packaged in a C file, and uploaded to the micro-controller to be interfaced by the TinyML library. Fig.~\ref{fig:cv_test} shows the performance of the computer vision model running with actual camera images. As shown, the real-life accuracy is on par with training accuracy, 96\%. The 4 output classes each corresponds to a vibration mode under a specific frequency, steering the robot to align with the target, or in the case of not finding a star, rotating the robot in place to scout the surroundings.

\section{Closed-Loop Autonomous Object Tracking}\label{sec:close-loop}

The integration of vision system (Section \ref{sec:vision}), supported by the on-board camera and pre-trained model, and motion system with frequency-based steering (Section \ref{sec:motion}) enables the autonomous operation of FAVbot. Open-source libraries are used for capturing and un-compressing images from the camera (Adafruit\_VC0706, JPEGDecoder), executing the CNN model (EloquentTinyML), outputting actuation signal (nRF52\_MBED\_PWM), and Bluetooth operations (ArduinoBLE).

Fig.~\ref{fig:software_flow_diagram} provides an overview of the robot software detailing both the characterization pipeline to determine frequency modes of the motion system and the autonomous pipeline using computer vision. The functionalities and each actuation mode's registered information (which consists of the driving frequency and driving duration for each actuation cycle) could be controlled or modified wirelessly through Bluetooth communication to minimize disturbance to the system for steering consistency. 

\new{Integer commands ranging from 1 to 100 set the actuation frequency in kilohertz accordingly which allows sweeping the complete frequency range in one characterization session.} Encoded command 103-106 will register the current frequency to the four motion modes corresponding to the detected target location: STRAIGHT, LEFT, RIGHT, and SEARCH, respectively. Command 203-206 enables the receiving of actuation cycle duration desired for each mode in multiples of 100 ms, e.g., a later command of 5 will set the duration for a single actuation cycle to 500 milliseconds. The purpose of programmable actuation duration is to balance off linear and angular speed differences among the four operation modes. Once manual characterization of the frequency range of interest is completed and the frequency and duration for each mode are registered, command 101 will change robot's operation to the autonomous mode (red shaded area in Fig.~\ref{fig:software_flow_diagram}).

\begin{figure*}[t!]
    \centering
       \includegraphics[width=\linewidth]{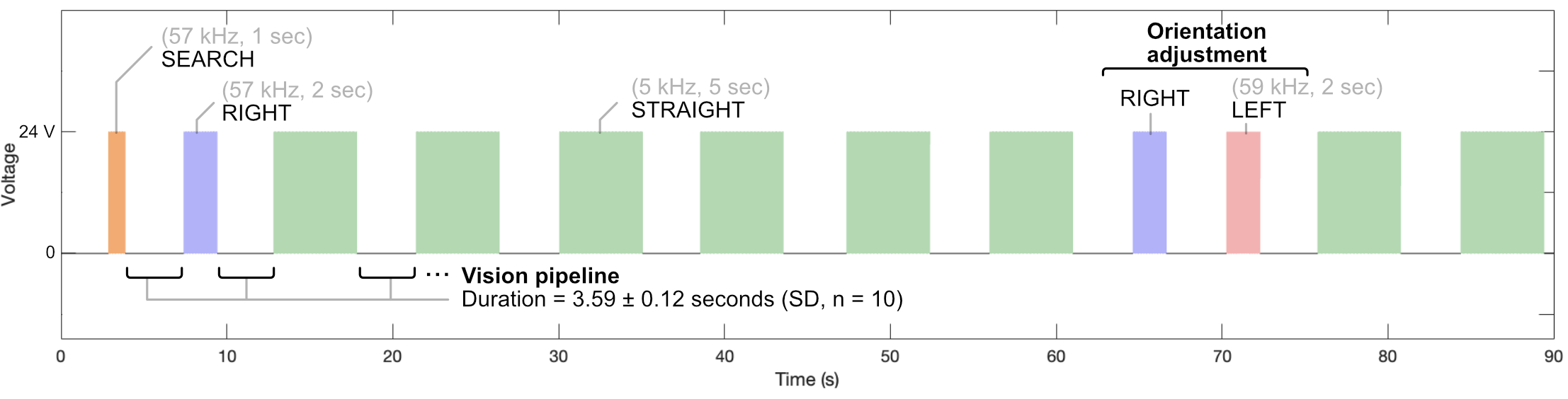}
       \caption{Reconstructed actuation segment corresponding to experiment in Fig.~\ref{fig:single_star_tracking_second_set}.}
       \label{fig:actuation_signal}
\end{figure*}

\begin{figure*}[t!]
    \centering
    \subfloat[\label{fig:single_star_tracking}]{%
       \includegraphics[width=0.5\linewidth]{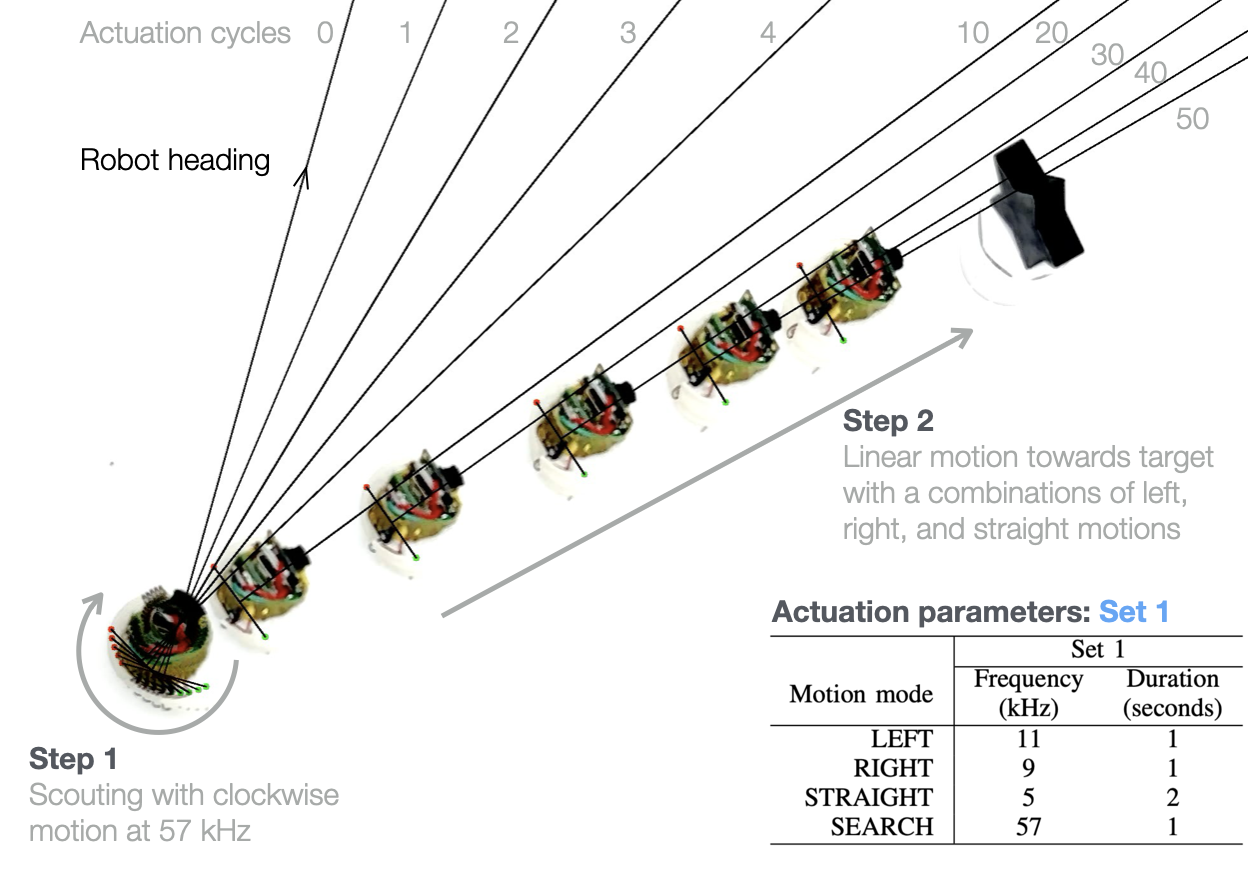}}
    \hfill
    \subfloat[\label{fig:single_star_tracking_second_set}]{%
       \includegraphics[width=0.5\linewidth]{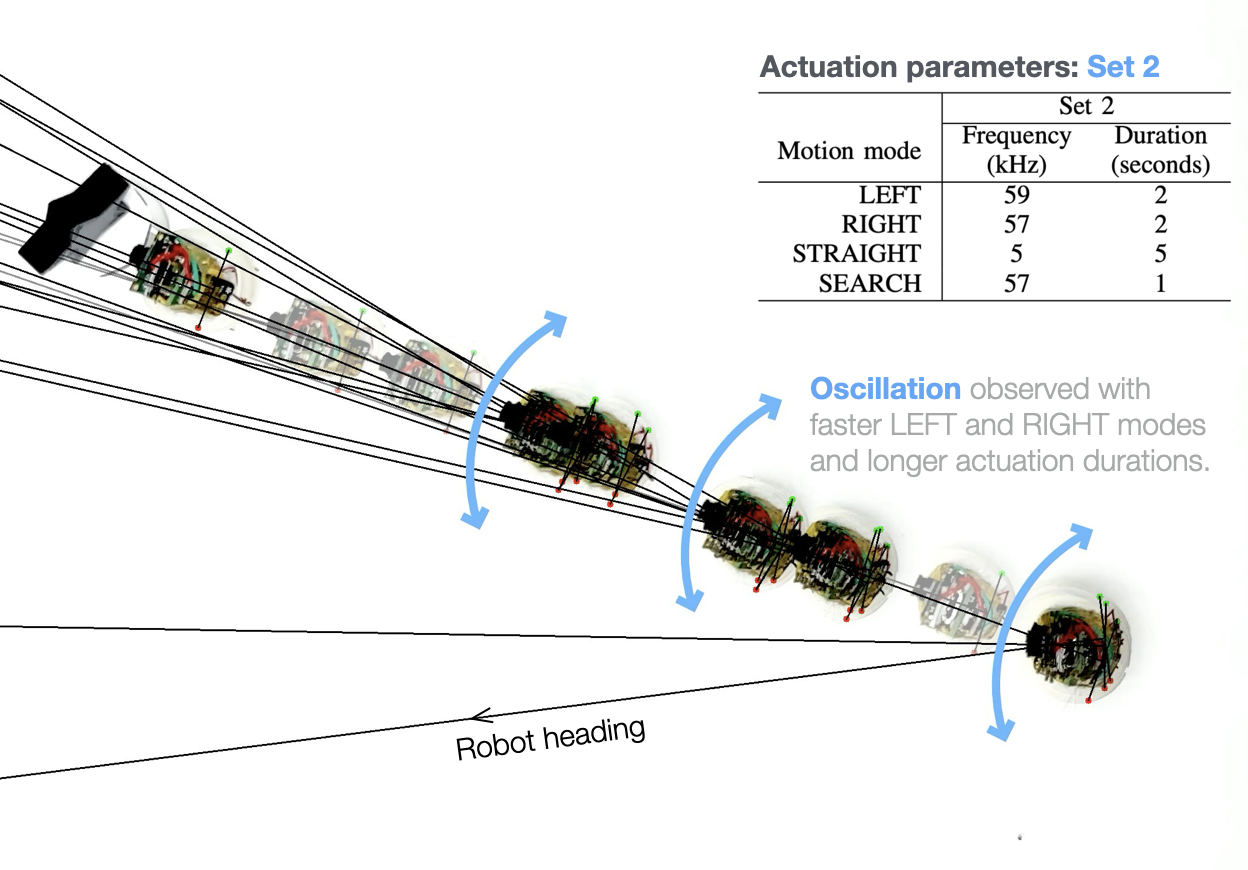}} \\
    \subfloat[\label{fig:multi_star_tracking}]{%
       \includegraphics[width=1.0\linewidth]{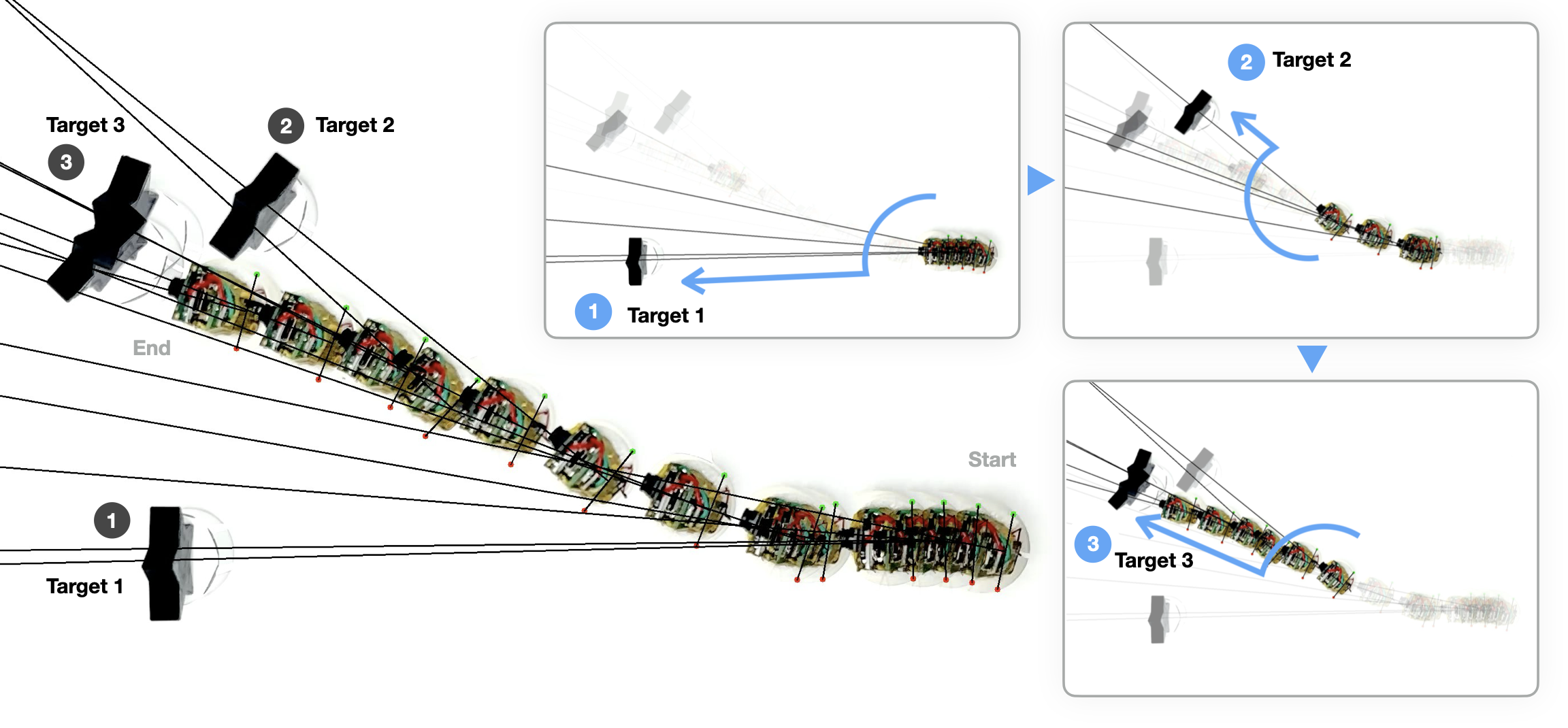}}
       
       \caption{Experimental results of FAVbot tracking object under different conditions. (a) Using actuation frequency set 1 in Table~\ref{table:actuation_parameters} to track a static star. (b) Using a different actuation frequency set 2 in Table~\ref{table:actuation_parameters} to track a static star. (c) Tracking of a moving target at 3 different locations. Videos of the experiments can be found in the multimedia attachment.}
       \label{fig:closed_loop_group}
\end{figure*}

In the autonomous mode, the robot will loop through image capturing, perform vision model inference to determine the relative location of the target, \new{and actuate the robot with previously registered frequency as characterized in Fig.~\ref{fig:motion_panel}} and the set duration to approach the target. For example, if the vision model predicts that the target is to the left of the robot's view, the LEFT mode will be actuated to correct the course. 

Fig.~\ref{fig:actuation_signal} shows a sequence of reconstructed actuation signal in an object tracking session, \new{during which the robot uses a combination of RIGHT, LEFT, and STRAIGHT modes to adjust its orientation to track the target autonomously.} Square waves with a peak voltage of 24 volts at various frequencies are associated with the four motion modes. Between the actuation signals is idle time for the vision pipeline, which takes 3.59 $\pm$ 0.12 seconds to capture and analyze the image. \new{During the idle time, the vibration is turned off to ensure the quality of imaging.}

\new{The tracking session is visualized in Fig.~\ref{fig:single_star_tracking}, where images captured from the top are overlaid to show the trajectory of the robot.} The images are augmented in post-analysis, where the two tracking markers are highlighted  with red and green dots and the changing orientations of the robot are computed and plotted with black lines. The actuation parameters used for this experiment are listed in Set 1, Table~\ref{table:actuation_parameters}. At the initial position, the robot is unable to detect the star target in the picture frame, thus in the first 4 actuation cycles, the robot executed the SEARCH mode and rotated clockwise. Once the star is found, the robot uses a combination of the other three modes to move towards the target.

\begin{table}[h!]
\centering
\caption{Actuation parameter sets}
\label{table:actuation_parameters}
\begin{tabular}{r|cc|cc}
\hline
            & \multicolumn{2}{c|}{Set 1 (Fig.~\ref{fig:single_star_tracking})}                                                                                    & \multicolumn{2}{c}{Set 2 (Fig.~\ref{fig:single_star_tracking_second_set})}                                                                                     \\ \cline{2-5} 
Motion mode & \begin{tabular}[c]{@{}c@{}}Frequency\\ (kHz)\end{tabular} & \begin{tabular}[c]{@{}c@{}}Duration\\ (seconds)\end{tabular} & \begin{tabular}[c]{@{}c@{}}Frequency\\ (kHz)\end{tabular} & \begin{tabular}[c]{@{}c@{}}Duration\\ (seconds)\end{tabular} \\ \hline
LEFT        & 11                                                        & 1                                                            & 59                                                        & 2                                                            \\
RIGHT       & 9                                                         & 1                                                            & 57                                                        & 2                                                            \\
STRAIGHT    & 5                                                         & 2                                                            & 5                                                         & 5                                                            \\
SEARCH  & 57                                                        & 1                                                            & 57                                                        & 1                                                            \\ \hline
\end{tabular}
\end{table}

To further demonstrate the robot's flexibility with resonance modes, a second experiment with a different set of actuation parameters (Set 2, Table~\ref{table:actuation_parameters}) is conducted and visualized in Fig.~\ref{fig:single_star_tracking_second_set}. In this experiment, the LEFT and RIGHT modes are registered with ``stronger'' rotation modes at 59~kHz and 57~kHz, respectively. These two modes offer more than double the angular velocities as used previously. Over-correction is thus observed in the experiment as shown with the curved arrows in the figure. This implies that with the current actuation and sensing architecture, the motion characteristics indicate an optimal actuation duration. To achieve an accurate trajectory, each actuation duration should be minimized for frequent vision feedback; However, the total mission duration will be elongated. On the contrary, longer actuation duration could lead to faster approaching to the target, but also has risks including oscillation or missed detection of the target.

To further test the adaptability of the robot to a dynamic environment with a moving target. An experiment during which the target is repositioned at different locations is visualized in Fig.~\ref{fig:multi_star_tracking}. The complete trajectory is decomposed into 3 segments for 3 selected target locations, shown in figure insets. The robot demonstrated effective vision and motion systems to continuously adjust robot's orientation and approach the updated target location.

\section{Conclusion}
In this paper, we presented a 3-cm miniaturized robot that is capable of autonomously tracking targets using frequency-controlled steering. Despite the constraints imposed by the limited volume, the robot is equipped with actuation, sensing, computation, power, and communication functionalities. These are accompanied by CNN-based computer vision that combines multiple circuit components to develop a control system that achieves full autonomy, marking it as the first of its kind in miniaturized robots.

With its novel actuation mechanism, FAVbot utilizes mechanical resonances induced by asymmetrical design for steering, achieving well-controlled motion in vibration-driven MMRs using a single actuator.
This new actuation mechanism overcomes the limitation. Multiple modes are observed across a wide spectrum of frequency from 1 to 62~kHz, each offering unique combinations of linear velocity, angular speed, direction of motion, and radius of curvature. Under certain frequencies, the robot is capable to turn clockwise and counterclockwise with near zero radius of curvature. Such modes are invaluable in need of scouting the 360$^\circ$ surroundings. This rich set of dynamics result from this actuation mechanism makes the robot a good candidate for many closed-loop applications including object search and tracking as demonstrated in this paper. 

Another key advantage of adopting the proposed frequency-controlled steering mechanism is the reduction of components required for actuation. In contrast to most wheeled robots whose turning is achieved by differential driving of two or more motors with accompanying driving electronics, our robot is able to achieve the same with a single low-profile piezoelectric actuator. In the pursuit of miniaturization, such simplification is highly advantageous. In addition, piezoelectric material as vibration source is also easier to scale down in comparison to motors, leaving space for other important integration at the small scale. It is worth noting that the current assembly relies on COTS components, which are not fully optimized in size. Yet FAVbot achieved a compact assembly in 3 cm, comparable to the smallest MMRs reported thus far. Future work involves using advanced integration and packaging techniques, such as thin-film piezoelectric materials, thin-film battery, and custom IC and CMOS sensors, and we expect the design to be scaled down by 3 times.

While the novel actuation mechanism brings advantages, it also posts new challenges -- one being the aforementioned reliability issue associated with vibration actuation that we still observed in some modes. The robot is sensitive to the change of terrine or existence of small obstacles at the bristle tips. This further emphasizes the value of having the computer vision closed-loop feedback to adjust the course in real-time. The actuation and vision systems together offer a robust solution. 

\new{Focusing on the closed-loop autonomy, the current robot uses commercially available small micro-controller with peripherals to reliable control the camera and actuator. Such system faces known challenges in power efficiency and memory availability, where we utilized a minimal-sized CNN model and achieved 15 minutes of untethered operation with a 40~mAh battery.}
For future development, we plan to make the robot more competitive in real-life applications in the directions of \new{improved power efficiency and memory allocation with a custom resistive RAM-based ASIC developed in conjunction~\cite{spetalnick202430}. In addition, the adoption of a custom ASIC further promises a more compact assembly, better efficiency of the vision pipeline, and improved autonomy with multi-target tracking and self-calibrating algorithms.}

\section*{Acknowledgment}

The authors would like to thank the GVU Prototyping Lab at Georgia Institute of Technology for providing high-resolution 3D printing capability. The authors would also like to thank Yixiao Hu, Yifan Shi and Tony Wang for the advice during the development.

\ifCLASSOPTIONcaptionsoff
  \newpage
\fi

\bibliographystyle{IEEEtran}

\bibliography{bibtex/bib/IEEEexample}

\begin{IEEEbiography}[{\includegraphics[width=1in,clip,keepaspectratio]{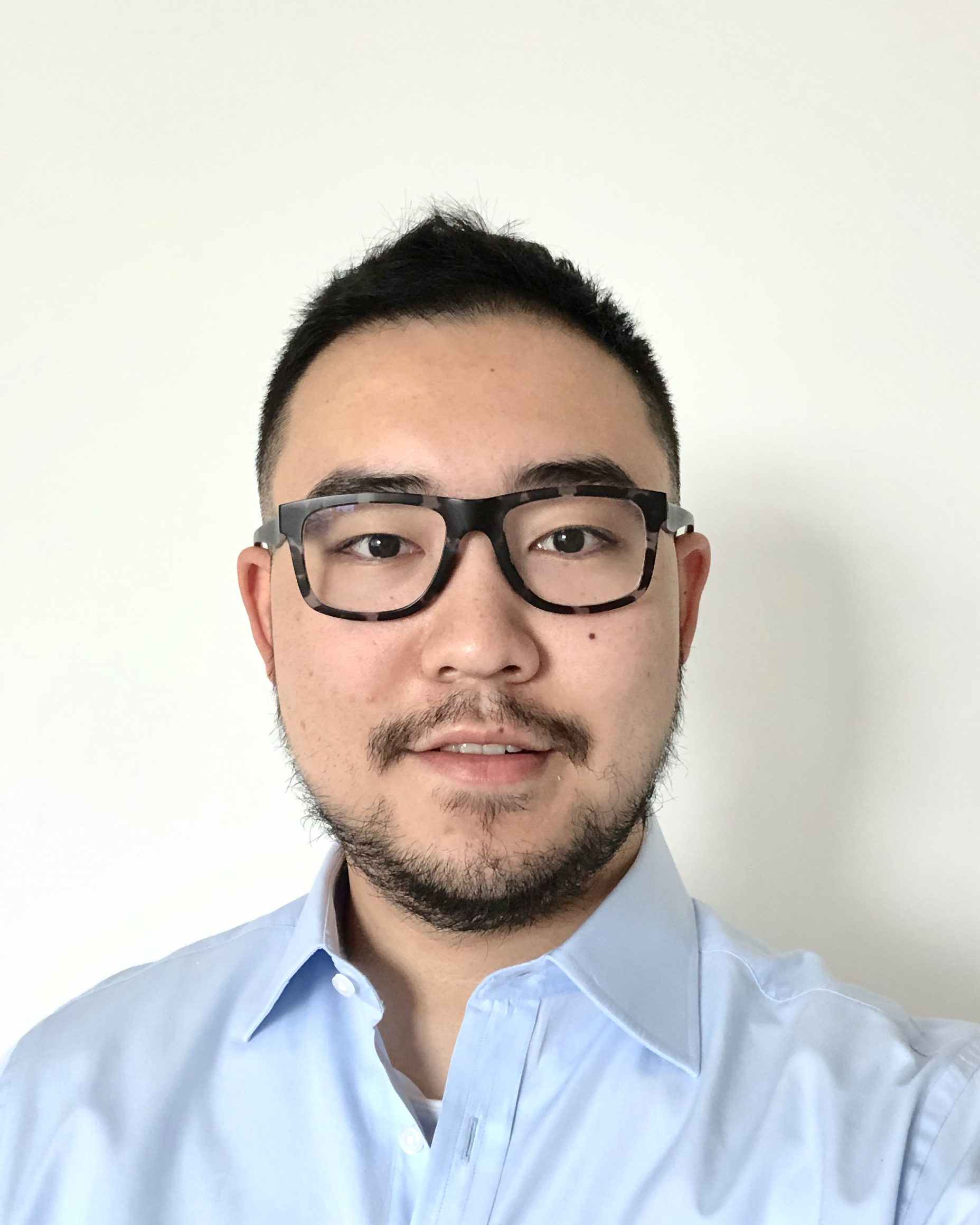}}]{Zhijian Hao}
(Student Member, IEEE) received the B.S.E. degree in electrical engineering from University of Michigan, Ann Arbor, MI, USA, in 2018, the M.S. degree in electrical and computer engineering from Georgia Institute of Technology, Atlanta, GA, USA, in 2020, and the Ph.D. degree in electrical and computer engineering from Georgia Institute of Technology, Atlanta, GA, USA, in 2024. His research focused on micro-robotics and sensor technology.
\end{IEEEbiography}

\begin{IEEEbiography}[{\includegraphics[width=1in,height=1.25in,clip,keepaspectratio]{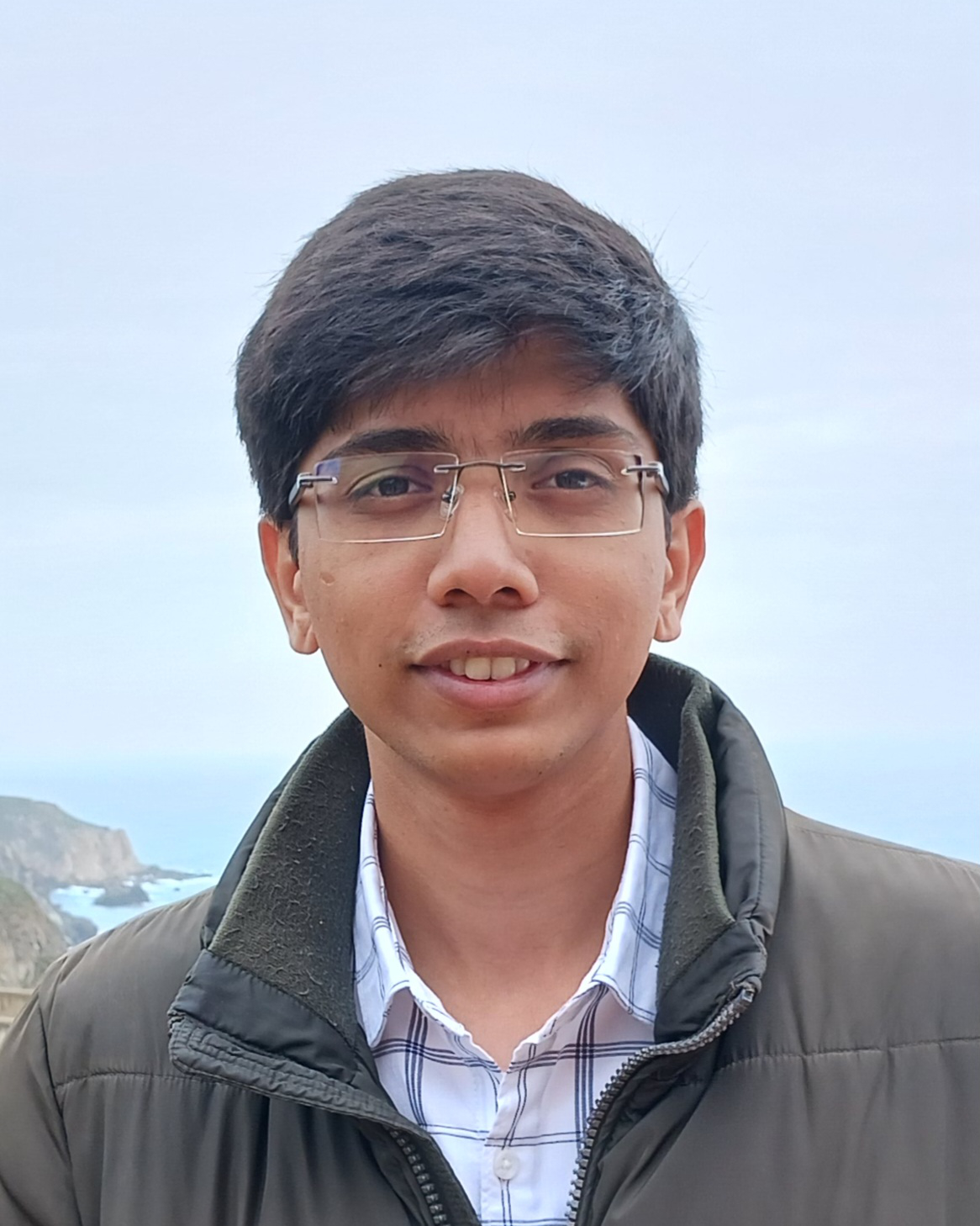}}]{Ashwin Lele} (Member IEEE) received the B.Tech. and M.Tech. degrees in electrical engineering from IIT Bombay, Mumbai, India, in 2019 and Ph.D. degree from the School of Electrical and Computer Engineering, Georgia Institute of Technology, Atlanta, USA in 2023. He previously held internship positions at Intel, Bangalore, India and Qualcomm Inc, San Jose, USA. He is currently a Principal Engineer with the Corporate Research Department, Taiwan Semiconductor Manufacturing Company (TSMC), San Jose, USA. His research interests broadly lie in low-power hardware design, emerging memory systems and their applications.
\end{IEEEbiography}

\begin{IEEEbiography}[{\includegraphics[width=1in,height=1.25in,clip,keepaspectratio]{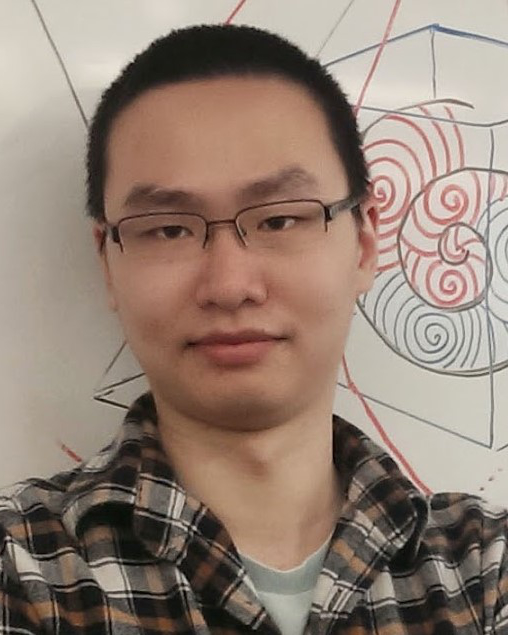}}]{Yan Fang} (Member, IEEE) received the B.S. degree in integrated circuits design and system integration from Xidian University, Xi’an, China, in 2010, and the M.S. and Ph.D. degrees in electrical engineering from the University of Pittsburgh, Pittsburgh, PA, USA, in 2013 and 2018, respectively.

He is currently an assistant professor at the Department of Electrical and Computer Engineering, Kennesaw State University. He was a postdoctoral researcher at Georgia Institute of Technology from 2018 to 2021. His research interests include neuromorphic computing, energy-efficient AI on edge and novel computing systems.
\end{IEEEbiography}

\begin{IEEEbiography}[{\includegraphics[width=1in,clip,keepaspectratio]{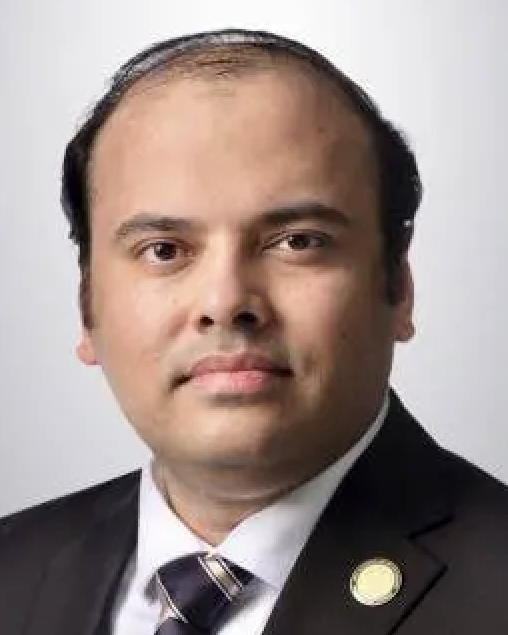}}]{Arijit Raychowdhury}
(Fellow, IEEE) received the Ph.D. degree in electrical and computer engineering from Purdue University, West Lafayette, IN, USA, in 2007. He joined the Georgia Institute of Technology (Georgia Tech), Atlanta, GA, USA, in January 2013. From 2013 to July 2019, he was an Associate Professor and held the ON Semiconductor Junior Professorship in the department. Prior to joining Georgia Tech, he held research positions at Intel Corporation for six years and Texas Instruments for one and a half years. He is currently the Steve W Chaddick Chair and a Professor with the School of Electrical and Computer Engineering, Georgia Tech. He is also the Director of the Center for the Co-Design of Cognitive Systems (CoCoSys), a Joint University Microelectronics Program 2.0. His research interests include low-power digital and mixed-signal circuit design, design of power converters, signal processors, and exploring interactions of circuits with device technologies. He holds more than 27 U.S. and international patents and has published over 300 papers in journals and refereed conferences. Dr. Raychowdhury is the winner of several prestigious awards, including the Purdue University Outstanding ECE Alumni Award 2023, SRC Technical Excellence Award in 2021, the Qualcomm Faculty Award in 2021 and 2020, the IEEE/ACM Innovator under 40 Award, the NSF CISE Research Initiation Initiative Award (CRII) in 2015, the Intel Labs Technical Contribution Award in 2011, the Dimitris N. Chorafas Award for outstanding doctoral research and best thesis in 2007, and several fellowships. He and his students have won 18 best paper awards over the years.
\end{IEEEbiography}

\begin{IEEEbiography}[{\includegraphics[width=1in,height=1.25in,clip,keepaspectratio]{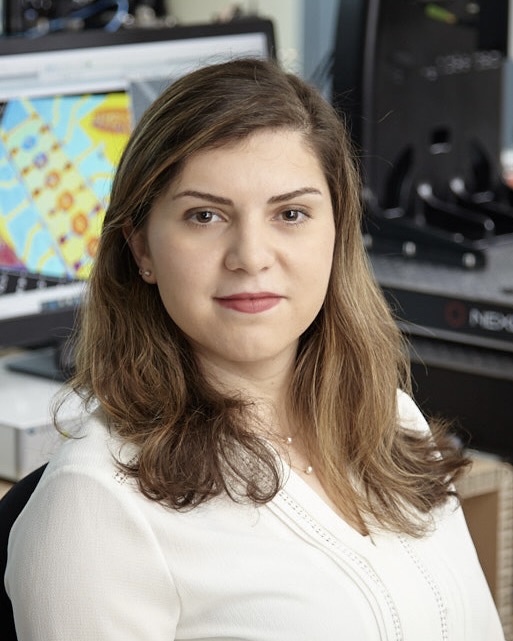}}]{Azadeh Ansari}
(Member, IEEE) received the B.S. degree in electrical engineering from the Sharif University of Technology, Tehran, Iran, in 2010, and the M.S. and Ph.D. degrees in electrical engineering from the University of Michigan, Ann Arbor, in 2013 and 2016, respectively. She is currently the Sutterfield family Assistant Professor with the School of Electrical and Computer Engineering, Georgia Institute of Technology. Prior to joining the ECE Faculty, she was a Post-Doctoral Scholar with the Department of Physics, California Institute of Technology. Her research interests cover nano/micro electromechanical systems (N/MEMS) and microscale robotics. She is the recipient of the 2021 Roger Webb Outstanding Junior Faculty Award at Georgia Tech, the 2020 NSF CAREER Award and the 2017 ProQuest Distinguished Dissertation Award from the University of Michigan.
\end{IEEEbiography}

\end{document}